\documentclass[lettersize,journal]{IEEEtran}
\usepackage{amsmath,amsfonts}
\usepackage{algorithm}
\usepackage{array}
\usepackage[caption=false,font=normalsize,labelfont=sf,textfont=sf]{subfig}
\usepackage{textcomp}
\usepackage{stfloats}
\usepackage{url}
\usepackage{verbatim}
\usepackage{graphicx}
\usepackage{cite}
\usepackage{xcolor}
\usepackage{multirow}
\usepackage{booktabs}
\usepackage{bbm}
\usepackage{mathtools}
\usepackage[normalem]{ulem}
\usepackage{xspace}
\usepackage{colortbl}
\usepackage[pagebackref,breaklinks,colorlinks,citecolor=blue,linkcolor=blue,bookmarks=false]{hyperref}
\usepackage{algpseudocode}

\newcommand{\calA}{{\mathcal{A}}}
\newcommand{\calD}{{\mathcal{D}}}

\newcommand{\calF}{{\mathcal{F}}}

\newcommand{\calL}{{\mathcal{L}}}
\newcommand{\calS}{{\mathcal{S}}}

\newcommand{\calX}{{\mathcal{X}}}

\newcommand{\Eref}[1]{Eq.~(\ref{#1})}
\newcommand{\Fref}[1]{Fig.~\ref{#1}}
\newcommand{\Tref}[1]{Table~\ref{#1}}
\newcommand{\Aref}[1]{Algorithm~\ref{#1}}

\newcommand{\norm}[1]{\| #1 \|}
\newcommand{\argmin}{\operatornamewithlimits{\arg \min}}

\def\onedot{.\@\xspace}
\def\eg{\emph{e.g}\onedot} 
\def\ie{\emph{i.e}\onedot}

\def\etal{\emph{et al}\onedot}

\DeclarePairedDelimiter\ceil{\lceil}{\rceil}

\begin{document}

\title{An Iterative Method for Unsupervised Robust Anomaly Detection under Data Contamination}

\author{
Minkyung Kim$^{1}$,
Jongmin Yu$^{2}$, 
Junsik Kim$^{3\dagger}$,
Tae-Hyun~Oh$^{4}$,
Jun~Kyun~Choi$^{1}$~\IEEEmembership{Senior Member,~IEEE}

\thanks{$^{1}$ School of Electrical Engineering, KAIST, Daejeon 34141, Republic of Korea; \{mkkim1778@kaist.ac.kr, jkchoi59@kaist.edu\}}
\thanks{$^{2}$ Department of Engineering, King's College London, Strand, London WC2R 2LS, England; \{jongmin.yu@kcl.ac.uk\}}
\thanks{$^{3}$ School of Engineering and Applied Sciences, Harvard University, Cambridge, MA 02138, U.S.A.; \{mibastro@gmail.com\}}
\thanks{$^{4}$ Dept. of Electrical Engineering and Graduate School of AI (GSAI), POSTECH, Pohang 37673, Republic of Korea, and Institute for Convergence Research and Education in Advanced Technology, Yonsei University, Seoul, Republic of Korea; \{taehyun@postech.ac.kr\}}
\thanks{$\dagger$ represents the corresponding author.}
}

\markboth{Journal of \LaTeX\ Class Files,~Vol.~14, No.~8, August~2021}%
{Shell \MakeLowercase{\textit{et al.}}: A Sample Article Using IEEEtran.cls for IEEE Journals}

\maketitle

\begin{abstract}
Most deep anomaly detection models are based on learning normality from datasets due to the difficulty of defining abnormality by its diverse and inconsistent nature.
Therefore, it has been a common practice to learn normality under the assumption that anomalous data are absent in a training dataset, which we call \textit{normality assumption}.
However, in practice, the normality assumption is often violated due to the nature of real data distributions that includes anomalous tails, \ie, \textit{a contaminated dataset}.
Thereby, the gap between the assumption and actual training data affects detrimentally in learning of an anomaly detection model.
In this work, we propose a learning framework to reduce this gap and achieve better normality representation.
Our key idea is to identify sample-wise normality and utilize it as an importance weight, which is updated iteratively during the training.
Our framework is designed to be model-agnostic and hyper-parameter insensitive so that it applies to a wide range of existing methods without careful parameter tuning.
We apply our framework to three different representative approaches of deep anomaly detection that are classified into one-class classification-, probabilistic model-, and reconstruction-based approaches.
In addition, we address the importance of a termination condition for iterative methods and propose a termination criterion inspired by the anomaly detection objective.
We validate that our framework improves the robustness of the anomaly detection models under different levels of contamination ratios on five anomaly detection benchmark datasets and two image datasets.
On various contaminated datasets, our framework improves the performance of three representative anomaly detection methods, measured by Area Under the ROC curve.
\end{abstract}


\section{Introduction}\label{sec:intro}
Anomaly detection is an identification problem of anomalous samples in a dataset.
It has crucial applications to prevent property loss or personal injury across various industries, such as credit card fraud detection~\cite{ahmed2016survey, hilal2021review}, predictive maintenance~\cite{carrasco2021anomaly, yang2019novel}, intrusion detection~\cite{marteau2021random, mothukuri2021federated}, and video surveillance~\cite{zhou2019anomalynet, huang2021abnormal, wang2021robust}.
If anomalies can be characterized in advance, an anomaly detection problem is boiled down to just a simple binary~\cite{abe2006outlier, amin2016radar, muhammad2018convolutional} or a multi-class classification~\cite{mabrouk2018abnormal} problem with an imbalanced dataset. 
However, in practice, it is challenging for even domain experts to determine what should be detected as abnormal cases in advance.
Therefore, most anomaly detection models have focused on learning a sense of normality from a dataset and identifying anomalous samples using the learned normality, where an anomalous sample is defined as an observation deviating considerably from normality~\cite{ruff2021unifying}.

By leveraging recent successes in deep learning, deep learning-based anomaly detection has been recently researched and has shown high capabilities~\cite{zhou2019anomalynet, huang2021abnormal, wang2021robust, muhammad2018convolutional, mabrouk2018abnormal, ruff2021unifying, chalapathy2018anomaly, ruff2018deep, an2015variational, schlegl2017unsupervised, perera2019ocgan, nachman2020anomaly, dias2020anomaly, rudolph2021same, gudovskiy2022cflow, sakurada2014anomaly, zhou2017anomaly, xia2015learning, beggel2019robust, fan2020robust, pang2020self, yu2021normality, zong2018deep, andrews2016detecting, erfani2016high, ergen2019unsupervised, golan2018deep, hendrycks2019using, bergman2020classification, lai2020robust}.

According to ways of modeling normality from a dataset, deep learning-based approaches are mainly categorized into: one-class classification-\cite{chalapathy2018anomaly, ruff2018deep}, probabilistic model-\cite{an2015variational, schlegl2017unsupervised, perera2019ocgan, nachman2020anomaly}, and reconstruction-based one~\cite{hinton2006reducing, masci2011stacked, sakurada2014anomaly, zhou2017anomaly}.
Each of them defines the concept of normality in its own way. 
Although these approaches train a model in an unsupervised way, they commonly assume all training data are normal, which we call \textit{normality assumption}.
The normality assumption in a dataset is hard to be held in practice, and the models designed with such assumption deteriorate performance rapidly when the ratio of anomalous data in a training dataset increases.
Thus, it is important for a practical anomaly detection model to be robust to a mixed dataset of normal and anomalous samples without labels. 
We call such a mixed set of normal and abnormal data as a contaminated dataset and the ratio of abnormal samples in the dataset as a contamination ratio.

Among deep learning-based anomaly detection, less research has been explored with a contaminated dataset.
The existing studies, \eg, ~\cite{xia2015learning, zhou2017anomaly, beggel2019robust, yu2021normality, pang2020self, fan2020robust, lai2020robust},
dealing with a contaminated training dataset mainly aim to reduce the negative effects from abnormal samples in the given dataset while learning the normality.
However, most of them rely on hyper-parameters that control which and how much data to consider as pseudo-normal/-abnormal samples. These vary depending on datasets and application scenarios. 
In addition, the majority of the existing studies focus only on reconstruction-based methods~\cite{hinton2006reducing, xia2015learning, zhou2017anomaly, beggel2019robust, yu2021normality} limiting their applications.
In contrast to the existing approaches, 
we break away from hyper-parameter-controlled pseudo-labeling 
and extend the applicability of the anomaly detection algorithm over the aforementioned three major categories.

In this work, we propose a robust anomaly detection learning framework to handle datasets with various contamination ratios.
To this end, we propose an iterative method to train a model
with sample-wise importance weights estimated from each training stage.
Our key idea is to identify the normality of each sample and utilize it as an importance weight.
The proposed learning method does not limit the base anomaly detection model type and it is free from hyper-parameters required for pseudo-labeling or the need for known contamination ratio information.
We also propose a 
termination criterion to measure the quality of the anomaly detection model during the iterative learning process.
Furthermore, we show the extension of our learning framework to encompass ensemble models that are known to achieve better robustness than single models. 
We empirically investigate the effect of contamination ratios on anomaly detection performance and analyze the effectiveness of the proposed framework on various benchmark settings: five anomaly detection benchmarks and two image datasets with different levels of contamination ratios.
On various datasets with contamination ratios from 31.6\% to 0.1\%, our framework improves the performance of three representative anomaly detection methods, measured by Area Under the ROC curve.
We summarize our main contribution as follows:
\begin{itemize}
  \item We propose an unsupervised deep anomaly detection framework robust to a wide range of contamination ratios, which is tailored for a model-agnostic and hyper-parameter insensitive learning framework.
  \item We propose a new termination criterion to measure the quality of an anomaly detection model based on the objective of anomaly scores.
  \item We demonstrate the effectiveness and generic applicability of our framework combined with three widely used anomaly detection approaches by 
   extensive analyses.
\end{itemize}

\section{Related Work}\label{sec:related works}
\subsection{Anomaly detection with normality assumption}\label{sec:related works-a}
According to the survey~\cite{ruff2021unifying}, the following three approaches are considered as the main groups of anomaly detection methods that are based on normality.
One-class classification-based approaches learn a discriminative decision boundary by mapping normal data to a compact representation. The most widely known shallow methods for this category are One-Class SVM (OC-SVM)~\cite{scholkopf2001estimating} and Support Vector Data Description (SVDD)~\cite{tax2004support}.
Later, OC-NN~\cite{chalapathy2018anomaly} and Deep SVDD~\cite{ruff2018deep} propose end-to-end deep learning approaches by replacing the shallow models with deep neural networks while using the same objective functions as in OC-SVM and SVDD.
They find a hyperplane and a minimum volume hypersphere,  respectively, for enclosing normal data in a latent space.
The recent work~\cite{maziarka2021oneflow} proposes a flow-based one-class classifier finding a minimal volume bounding region.
Another line of classification-based approaches~\cite{golan2018deep, hendrycks2019using, bergman2020classification} has been studied by leveraging self-supervised learning, mainly for image data. 
In these approaches, normality is measured by the error of surrogate tasks such as
augmentation classifications of rotation, flip, or patch re-arrangement.

Among the probabilistic model-based methods, kernel density estimation (KDE)~\cite{parzen1962estimation, scott2015multivariate} is the most widely used shallow non-parametric density estimator along with GMM~\cite{roberts1994probabilistic, zong2018deep}. 
Recently, anomaly detection studies applying generative models, such as Variational Autoencoder~\cite{kingma2013auto}, Generative Adversarial Networks (GAN)~\cite{schlegl2017unsupervised}, and Normalizing Flows~\cite{rezende2015variational}, have also been actively conducted~\cite{an2015variational, perera2019ocgan, nachman2020anomaly}, where
they learn a data 
distribution in a latent feature space.

In the reconstruction-based approaches, Autoencoder (AE)~\cite{hinton2006reducing, masci2011stacked} is the most commonly used method, and its 
various variants have been proposed for anomaly detection~\cite{sakurada2014anomaly, zhou2017anomaly}. 
In addition, there have been attempts to fuse 
the discriminative representation of AE with shallow methods~\cite{andrews2016detecting, erfani2016high, ergen2019unsupervised}.

Although there have been extensive studies for anomaly detection with normality modeling, the aforementioned approaches require a clean training dataset that does not contain abnormal data, which is hard to be held in practice.
It often results in sub-optimal performance and limits the applicability of the existing methods.

\subsection{Anomaly detection with contaminated dataset}\label{sec:related works-b}
While one-class classification-based approaches, such as OC-SVM, SVDD, and its deep variants, may deal with the contaminated dataset setting using a contamination ratio as a hyper-parameter, they require prior knowledge of the ratio in advance. 
However, the contamination ratio information is unknown in practice. 

Robust approaches~\cite{xia2015learning, zhou2017anomaly, beggel2019robust, yu2021normality, pang2020self, fan2020robust} have been proposed to tackle learning normality from contaminated datasets, which aim to reduce the negative effects of abnormal data mixed in a contaminated training dataset.
To this end, most of the studies exploit 
pseudo-labeling 
to improve the robustness of their models, where
pseudo-labeling is done particularly by AE~\cite{xia2015learning, zhou2017anomaly, beggel2019robust, yu2021normality} or 
regression~\cite{pang2020self, fan2020robust}.

In the early work by Xia~\etal\cite{xia2015learning}, pseudo-labeling is performed based on the reconstruction error of AE. 
Then, AE is trained iteratively to reduce the reconstruction error of pseudo-normal data with a regularization term representing the separability of the error distribution.
Zhou~\etal\cite{zhou2017anomaly} and Beggel~\etal\cite{beggel2019robust} follow the similar framework, but 
Zhou~\etal use robust AE inspired by Robust PCA~\cite{candes2011robust}, and Beggel~\etal use adversarial AE~\cite{makhzani2015adversarial}.
Yu~\etal\cite{yu2021normality} use a generative adversarial network (GAN) to generate high-confidence pseudo-normal samples and pseudo-label the data by measuring the similarity between each data and the generated samples. 

While most of the aforementioned works are based on AE and its variants, a few recent approaches propose to use regression models~\cite{pang2020self, fan2020robust} that are trained iteratively.
Pang~\etal\cite{pang2020self} and Shim~\etal\cite{shim2018high} introduce a two-class ordinal regression model with 
neural networks, and Fan~\etal\cite{fan2020robust} deploy a Gaussian process regression model. 
Both works require an initialization stage of anomaly detection performed by a separate off-the-shelf anomaly detector. 
Then, the model is iteratively trained on the pseudo-labeled data.

One important limitation of the aforementioned approaches is that they require 
hyper-parameters that control which and how much data should be considered as pseudo-normal or pseudo-abnormal samples.
However, optimal hyper-parameters may vary depending on datasets and unknown contamination ratios.
Unlike the previous approaches using hard pseudo-labels, which assign labels to normal or abnormal, 
pseudo-labels can be also continuous values in [0, 1], \ie, soft pseudo-label.
Similar to our work, early works~\cite{abe2006outlier, ide2016change} use soft pseudo-labels to prioritize samples differently in training, depending on notions of confidence. 
However, these methods are applied to different scopes of problems: normality assumption~\cite{abe2006outlier}, learning shallow models~\cite{abe2006outlier, ide2016change}, or time-series domain~\cite{ide2016change}.

Different from the approaches
relying on pseudo-labeling,
Lai~\etal\cite{lai2020robust} propose a robust AE model by projecting a latent feature into a lower-dimensional subspace with a linear projection layer based on the idea analogous to PCA~\cite{wold1987principal} to detect anomalies. 
It shows improved results over AE-based methods, but whether it can be applied to other types of anomaly detectors remain an open problem. 

In this work, we deviate from using subsets of pseudo-labeled data that are controlled by hyper-parameters and propose a versatile learning framework that can be applied to all three main categories of anomaly detection approaches.

\section{Preliminary}\label{sec:preliminary}
We provide a brief introduction to three major approaches and corresponding methods that are widely used for normality modeling following the unifying view~\cite{ruff2021unifying}.
We denote 
a learning model with a set of learnable weights $\theta$ as $f_\theta$, a sample loss function as $l$, a total loss as $\calL$, an anomaly score function as $s$, and a given dataset as $\calD=\{\boldsymbol x_1, \cdots, \boldsymbol x_n\}$, where $\boldsymbol x_i \in \mathbb{R}^d$,
and $\calL$ is the average of sample losses across all the data points as follows: 
\begin{align*}
    \calL = \frac{1}{n} \sum_{i=1}^{n} l(\boldsymbol x_i).
\end{align*}

\begin{figure*}[t]
\centering
\subfloat[Learning process of an anomaly detection model]{\includegraphics[width=0.9\linewidth]{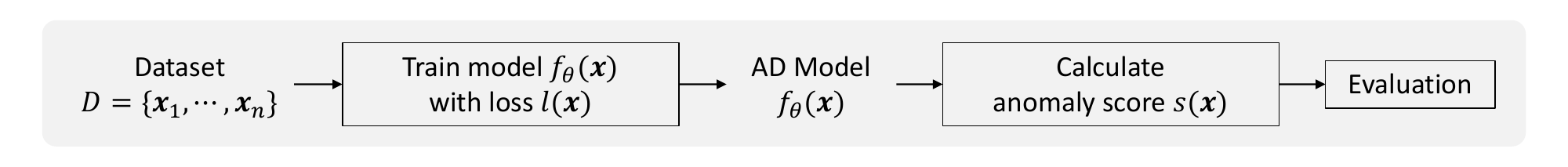}
\label{fig:framework-a}
}
\vfil
\subfloat[Learning process of an anomaly detection model applied to the proposed framework]{\includegraphics[width=0.9\linewidth]{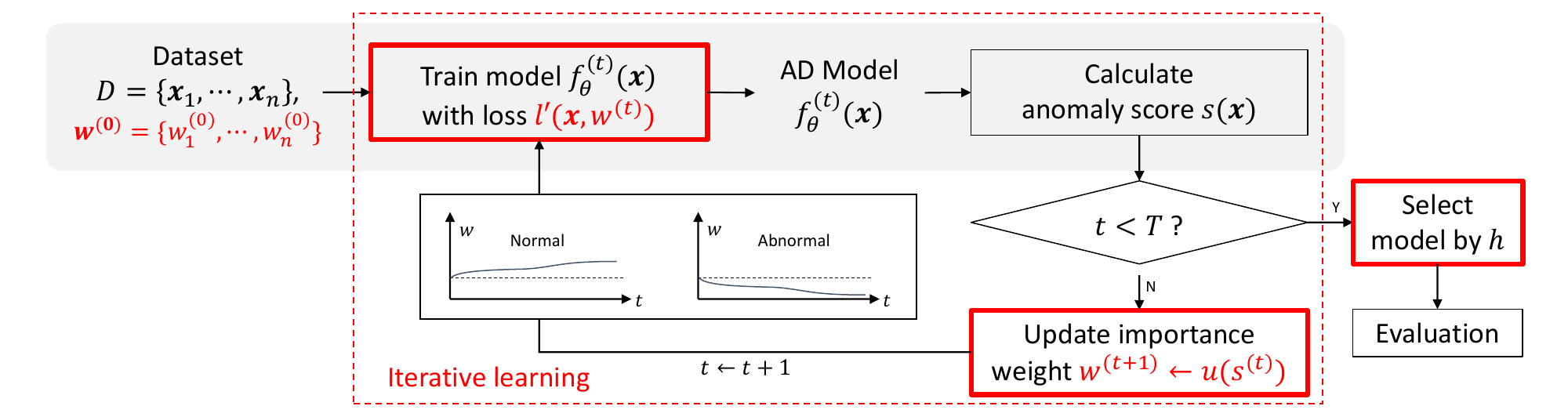}  \label{fig:framework-b}
}
\caption{(a) Learning process of existing anomaly detection models.
(b) Learning process of an anomaly detection model applied to the proposed framework
(Iterative Anomaly Detection; IAD). 
In IAD, both the importance weights $w$ of data samples and the model are refined and updated during iterative learning.
The model convergence is evaluated by the proposed termination criterion $h$.
The rectangle boxes 
marked in red indicate 
the main components of IAD, and the small graphs illustrate 
the expected changes in the importance weight of the true normal samples and the true abnormal samples in the contaminated dataset during the model training.}
\label{fig:framework}
\end{figure*}

We define an anomaly score function as $s$ that is measured by using a pre-trained normality model $f_\theta$. 
Then, anomaly detection is conducted by thresholding an anomaly score for each data.
However, since setting the threshold depends on a dataset or its application,
the performance of anomaly detection is measured with various threshold settings.
The series of these processes are shown in \Fref{fig:framework-a}.

\subsection{One-class classification-based: Deep SVDD}
Deep SVDD~\cite{ruff2018deep} is inspired by kernel-based SVDD and minimum volume estimation. Based on the capability of deep model learning good feature representation, Deep SVDD trains a neural network by minimizing the volume of a hypersphere that encloses the network representations of data. Deep SVDD maps data $\boldsymbol x\in\calX$ to a feature space $\calF$ through $\phi(\cdot;\theta): \calX \rightarrow \calF$, while gathering data around the center $\boldsymbol c$ of the hypersphere in the feature space, where $\theta$ denotes the set of weights of the neural network $\phi$. 
An anomaly score is measured by the distance between the center and a data point.

There are two versions of Deep SVDD. 
The \textit{soft-boundary} Deep SVDD explicitly optimizes the radius of the hypersphere with the hyper-parameter $\nu$, which controls the trade-off between the radius $R$ and the amounts of data points outside the hypersphere. 
Three modeling components of \textit{soft-boundary} Deep SVDD are as follows:
\begin{align}
\label{eqn:sb}
    f_\theta(\boldsymbol x_i) &:= \norm{\phi(\boldsymbol x_i; \theta) - \boldsymbol c}^2, \\
    l(\boldsymbol x_i) &:= \nu \cdot R^2 + \textrm{max}(0, f_\theta(\boldsymbol x_i)-R^2), \\
    s(\boldsymbol x_i) &:= f_\theta(\boldsymbol x_i).
\end{align}
For simplicity, we omit the regularizers in the rest of this section.

The other one, called \textit{One-Class} Deep SVDD, is designed explicitly under the normality assumption. 
Therefore, it penalizes the mean distance over all the data samples. 
Although it does not output a decision boundary, 
it is trained without data-dependent hyper-parameters. 
Three modeling components of \textit{One-Class} Deep SVDD are as follows:
\begin{align} 
\label{eqn:oc}
    f_\theta(\boldsymbol x_i) &:= \norm{\phi(\boldsymbol x_i; \theta) - \boldsymbol c}^2, \\
    l(\boldsymbol x_i) &= s(\boldsymbol x_i) := f_\theta(\boldsymbol x_i).
\end{align}

\subsection{Probabilistic-based: Normalizing flow}
Normalizing Flows (NF) is a family of generative models which model 
a likelihood
distribution
of given data samples. 
The merit of NF over other generative models is that it can calculate the likelihood of a data sample directly without any approximation gap. With this merit, NF has been applied in anomaly detection~\cite{nachman2020anomaly, dias2020anomaly, rudolph2021same, gudovskiy2022cflow}.

NF builds a normalizing flow between a base distribution $p_Z:\mathbb{R}^d \rightarrow \mathbb{R}$ and the target distribution $p_X:\mathbb{R}^d \rightarrow \mathbb{R}$ for a given dataset. 
By the change of variable formula, $p_X(\boldsymbol x)$ can be represented as:
\begin{align*}
    p_X(\boldsymbol x) = p_Z(g^{-1}(\boldsymbol x))|\det\mathbf{J}_{g^{-1}}(\boldsymbol x)|,
\end{align*}
where a base distribution $p_Z$ is a known and tractable probability density function for a random variable $Z$, $g$ denotes a random variable transform as $\boldsymbol x=g(\boldsymbol z)$, $g^{-1}$ is its inversion, and
$\mathbf{J}_{g^{-1}}$ is the Jacobian of $g^{-1}$. 
It is formally proven that if $g$ can be arbitrarily flexible, one can generate any distribution $p_X$ from a base distribution $p_Z$ (See~\cite{kobyzev2021normalizing} for more details). 
Without loss of generality, we denote $g$ as a deep network $\phi$ with its parameter $\theta$, which is usually trained by maximum likelihood estimation.
An anomaly score is measured by the negative likelihood from the trained NF. Three modeling components of NF are as follows:
\begin{align} 
\label{eqn:nf}
    f_\theta(\boldsymbol x_i) &:= p_Z(\phi^{-1}(\boldsymbol x_i; \theta))|\det\mathbf{J}_{\phi^{-1}_{\theta}}(\boldsymbol x_i)|, \\
    l(\boldsymbol x_i) &:= -\log(f_\theta(\boldsymbol x_i)), \\
    s(\boldsymbol x_i) &:= -f_\theta(\boldsymbol x_i).
\end{align}

\subsection{Reconstruction-based: Autoencoder}
Deep Autoencoder (AE) is a neural network used to learn representation by self-reconstruction, typically for dimension reduction or feature extraction. 
AE consists of two parts; an encoder $\phi$ and a decoder $\psi$. For a given dataset, the encoder maps data $\boldsymbol x \in \calX$ to a low-dimensional feature space $\calF$ through $\phi(\cdot;\theta_\phi): \calX \rightarrow \calF$, and the decoder maps the representation back to the reconstruction of the data through $\psi(\cdot;\theta_\psi): \calF \rightarrow \calX$, where $\theta_\phi$ and $\theta_\psi$ denotes the set of weights of the encoder and decoder networks, respectively. 
In the previous studies~\cite{sakurada2014anomaly, xia2015learning}, AE is shown to be capable of detecting anomalies.
Since AE trained with the normality assumption is expected to reconstruct normal data accurately but not do so with abnormal data, the reconstruction error is used as an anomaly score. Three modeling components of AE are as follows:
\begin{align} 
\label{eqn:ae}
    f_\theta(\boldsymbol x_i) &:= \psi(\phi(\boldsymbol x_i; \theta_\phi); \theta_\psi), \\
    l(\boldsymbol x_i) &= s(\boldsymbol x_i) := \norm{\boldsymbol x_i - f_\theta(\boldsymbol x_i)}_2^2.
\end{align}

\section{Iterative Anomaly Detection}\label{sec:method}\label{sec:method:overview}

When a dataset is contaminated, naively training anomaly detectors with the normality assumption would end up in a suboptimal solution.
In our proposed framework, which is named Iterative Anomaly Detection (IAD), the normality of each sample is estimated and used as an importance weight during training. 
A simple but effective way to estimate the normality of each sample is to utilize an anomaly score obtained from a base anomaly detection model that is applied to our framework.
In our IAD, a base anomaly detection model $f_\theta$ is iteratively trained with the proposed re-weighted sample loss function $l'(\boldsymbol x, w)$, 
where $w$ denotes an importance weight of data $\boldsymbol x$. 
Our IAD framework is summarized in~\Fref{fig:framework-b}.

The learning process consists of inner- and outer-loops. 
We term ``iteration'' for indicating inner-loop iterations of training 
a model $f_\theta$ given fixed importance weights,
and ``round'' for indicating outer-loop iterations (see the dotted red box in~\Fref{fig:framework-b}).

\subsection{Discriminative learning with importance weights}\label{sec:method:iad}
Without properly dealing with contaminated datasets, 
abnormal data samples
deteriorate the representation of normality in the training phase. 
To alleviate the susceptibility, we introduce a sample-wise importance weight variable $w$ into a loss function $l$ to learn normality discriminatively. 
Compared to the existing approaches using pseudo-labeling, we take a weight-based approach to reduce the artifact of the mis-classification in pseudo-labeling that may act as another source of outliers.
We indicate a model trained at round $t \in \{0, 1, \cdots\}$ as $f_\theta^{(t)}$, and 
define the total loss
as follows:
\begin{equation}
\label{eqn:dl}
    \calL=\frac{1}{n}\sum\nolimits_{i=1}^{n} l'(\boldsymbol x_i, w_i^{(t)}) 
             =\frac{1}{n}\sum\nolimits_{i=1}^{n} w_i^{(t)} \cdot l(\boldsymbol x_i),
\end{equation}
where $w_i^{(t)}$ is an importance weight updated at round $t$.
Since we do not have information of importance weights at the initial round
$t=0$, we set all the initial weights by $w_i^{(0)} = 1$.
Therefore, $f_\theta^{(0)}$ is equal to the base model, $f$.

\subsubsection{Importance weight update}
\begin{figure}[!t]
\centering
\includegraphics[width=2.5in]{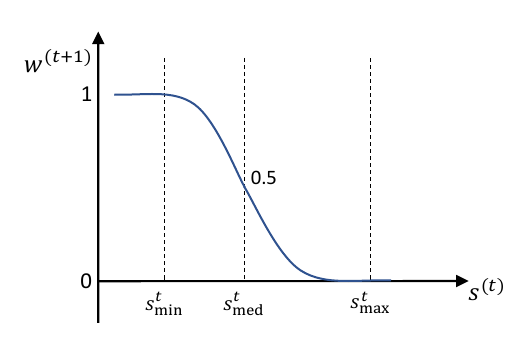}
\caption{Shape of the proposed importance weight update function $u$.}
\label{fig:update}
\end{figure}

Importance weights should assign 
a stronger learning signal when a sample is normal (expect to have a low anomaly score), while it should suppress an effect 
from an abnormal sample (expect to have a high anomaly score). 
To this end, we propose an importance weight update function $u$ to obtain $w_i^{(t+1)}$ as:
\begin{equation}
\label{eqn:update}
    w_i^{(t+1)} = u(s_i^{(t)}) = \frac{1}{1+e^{\alpha^{(t)}(s_i^{(t)}-\beta^{(t)})}},
\end{equation}
where $s_i^{(t)}$ denotes an anomaly score calculated by $s(\boldsymbol x_i)$ at round $t$.
The shape of $u$ is shown in \Fref{fig:update}.
The parameters of \Eref{eqn:update}, $\alpha^{(t)}$ and $\beta^{(t)}$, are obtained from
the statistics 
of anomaly scores;  
\ie, minimum, maximum, and median values of $\calS^{(t)}=\{s_1^{(t)},\cdots,s_n^{(t)}\}$ denoted as
$s^{(t)}_{\rm min}$, $s^{(t)}_{\rm max}$, and $s^{(t)}_{\rm med}$. 
Given the statistics, we define the parameters as follows: 
\begin{align}
\label{eqn:ab}
    \alpha^{(t)} &= \frac{1}{{\rm min}( s^{(t)}_{\rm med} - s^{(t)}_{\rm min},\ s^{(t)}_{\rm max} - s^{(t)}_{\rm med} )\cdot\tau}, \\
    \beta^{(t)} &= s^{(t)}_{\rm med}.
\end{align}
There are two motivations for $u$ design.
First, importance weights are increased (decreased) for samples with a lower (higher) anomaly score and vice versa.
Second, it aims to enforce the separability of the model during training by quickly adapting the importance weight to $1$ or $0$ according to the estimated normality. 
To this end, the shape of the sigmoid function is adjusted so that the data with a minimum (maximum) anomaly score has an importance weight of 1 (0).
$\tau>0$ is a scalar temperature parameter.
Since the importance weight $w$ is updated from the model trained in the previous round, iteratively training the model with the updated importance weights refines both the importance weights and the model.

\subsection{Termination criterion}                     
For any iterative process, an appropriate measure for termination is required. 
We propose a termination criterion based on two properties; convergence and grouping. 
When model parameters converge during training, the sample-wise normality is also expected to converge, \eg, the order of sample-wise anomaly scores no longer changes between the rounds. In practice, due to a deep model's stochastic training and high complexity, the order of anomaly scores frequently changes over rounds, even after a sufficient number of rounds. 
Although the changes in rankings may measure model convergence, 
measuring ranking changes alone is subtle or unstable
for anomaly detection models.
Nonetheless, we observe that the orders are locally flipped in most cases, but significant order flips rarely happen.
According to this observation, we develop our termination criterion to capture the convergence of our iterative learning.

Suppose a sorted samples from a dataset according to anomaly score.
Typically, anomaly of data is detected by thresholding the estimated anomaly scores, \ie, a binary labeling problem.
In the sorted sample list, local rank swaps 
do not affect performance unless changed normality or abnormality of the samples. 
The critical ranking changes happen only when a normal sample rank and an abnormal sample rank are swapped. 
We quantify this by grouping a dataset into two partitions; one with low anomaly scores and the other with high anomaly scores by percentile (we set 50\% for convenience). 
We measure the frequency of the changes in ranking across these two partitions as a termination criterion:
\begin{equation}
\label{eqn:termination}
    h^{(t)} = \sum_{i=1}^{n} {\mathbbm{1}_{\calA^{(t)}}(\boldsymbol x_i)},
\end{equation}

\begin{equation}
{\calA^{(t)}} = \left\{
              \boldsymbol x_i \;\middle|\;
              \begin{aligned}
              & \left( r_i^{(t)} < \ceil*{\tfrac{n}{2}} \wedge r_i^{(t-1)} > \ceil*{\tfrac{n}{2}}\right) \; \lor \; \\
              & \left(r_i^{(t)} > \ceil*{\tfrac{n}{2}} \wedge r_i^{(t-1)} < \ceil*{\tfrac{n}{2}}\right)
              \end{aligned}
            \right\},            
\end{equation}
where $h^{(t)}$ denotes the termination criterion value calculated at round
$t$, $r_i^{(t)}$ denotes the rank of $\boldsymbol x_i$ when anomaly scores at $t$ are sorted in ascending order, and $\calA^{(t)}$ denotes a set of samples that have changed their rank across the partitions, \eg the halfway point.
The termination metric $h$ counts the number of samples that belong to set $\calA^{(t)}$.
This value will decrease until there is no significant change in anomaly scores during iterative learning. 
However, since the proposed framework utilizes estimated importance weights during learning, there is no guarantee that this value will monotonically decrease.
Therefore, we propose to select the model with the lowest $h$ motivated by early termination~\cite{zhang2017understanding}.
We outline the learning process of IAD in \Aref{alg:iad}.

\begin{algorithm}[t]
\caption{Iterative Anomaly Detection}\label{alg:iad}
\begin{algorithmic}[1]
    \Require {Unlabeled dataset $\calD$, Base Model $f_\theta$, \newline \hspace*{4mm} Maximum number of rounds $T$, \newline \hspace*{4mm} Maximum number of epochs $E$}
    \Ensure $f_\theta^{(t)}$ 
    
    \Statex 
    \State  $t \leftarrow 0$
    \State  $w_i^{(0)} \leftarrow 1, i=\{1,\cdots,n\}$
    \For{$t=0$ to $T$} \Comment{outer-loop}
        \For{$e=1$ to $E$} \Comment{inner-loop}
            \State  Train $f_\theta^{(t)}$ with $\calL$
        \EndFor 
        \State  Calculate anomaly scores $s_i^{(t)}$ 
        \State  Calculate termination criterion value $h^{(t)}, t \ge 1$.
        \State  Update $w_i^{(t+1)} = u(s_i^{(t)}; \calS^{(t)})$. 
    \EndFor 
    
    \State \Return $\argmin_{t} h^{(t)}$
    
    \end{algorithmic}
\end{algorithm}

\section{Experiment}\label{sec:experiments}

\begin{table}[t]
\caption{Anomaly detection benchmarks.}\label{tab:dataset}
\centering
\begin{tabular}{lrrr}
        \toprule
        \multicolumn{1}{c}{Dataset} & \multicolumn{1}{c}{$n$} & \multicolumn{1}{c}{$d$} & \multicolumn{1}{c}{\# outliers (\%)} \\ \midrule
        Satellite   & 6,435     & 36    & 2,036 (31.6\%)    \\
        Arrhythmia  & 452       & 274   & 66 (14.6\%)       \\
        Cardio      & 1,831     & 21    & 176 (9.6\%)       \\
        Thyroid     & 3,772     & 6     & 93 (2.5\%)        \\
        Satimage-2  & 5,803     & 36    & 71 (1.2\%)        \\ \bottomrule
    \end{tabular} 
\end{table}

\begin{table*}[t]
\caption{Results on anomaly detection benchmarks. 
The contamination ratio is indicated below each dataset name.
We report the average AUC with a standard deviation computed over 10 seeds.
\textit{Base} denotes
each base anomaly detection model.
IAD represents the model selected by the proposed termination criterion, and 
IAD-Best represents the best model during iterative learning.
The results of IAD with the improvement compared to the base performance are indicated in bold.}\label{tab:table1}
\centering
\begin{tabular}{lllccccc}
\toprule
\multirow{2}{*}{Approach}                 & \multirow{2}{*}{Base model} & \multirow{2}{*}{Learning method} & Satellite & Arrhythmia & Cardio & Thyroid & Satimage-2 \\ \cmidrule(lr){4-8} 
                                          &                             &                                  & 31.6\%    & 14.6\%     & 9.6\%  & 2.5\%   & 1.2\%      \\ \midrule
\multirow{3}{*}{One-class classification} & \multirow{3}{*}{Deep SVDD} & Base                      & $68.2\pm5.8$          & $74.7\pm2.9$          & $64.5\pm5.4$          & $78.3\pm7.3$          & $86.7\pm6.8$          \\
                                          &                            & IAD                         & $\mathbf{80.7\pm4.0}$ & $\mathbf{80.0\pm2.4}$ & $\mathbf{85.3\pm5.1}$ & $\mathbf{86.7\pm3.1}$ & $\mathbf{97.6\pm2.1}$ \\ 
                                          &                            & IAD-Best                    & $82.4\pm2.5$          & $82.8\pm1.1$          & $88.1\pm4.8$          & $91.7\pm2.1$          & $99.1\pm0.3$          \\ \midrule 

\multirow{3}{*}{Probabilistic model}      & \multirow{3}{*}{MAF}       & Base                      & $72.3\pm3.1$          & $84.4\pm1.0$          & $86.9\pm2.5$          & $96.3\pm0.4$          & $99.7\pm0.4$          \\
                                          &                            & IAD                         & $\mathbf{74.3\pm2.2}$ & $\mathbf{85.9\pm0.8}$ & $\mathbf{90.3\pm2.9}$ & $\mathbf{97.3\pm0.2}$ & $\mathbf{99.8\pm0.3}$ \\  
                                          &                            & IAD-Best                    & $77.4\pm1.3$          & $88.4\pm0.6$          & $92.9\pm0.7$          & $97.8\pm0.1$          & $99.9\pm0.1$          \\ \midrule 

\multirow{3}{*}{Reconstruction}           & \multirow{3}{*}{AE}        & Base                      & $62.0\pm1.6$          & $74.0\pm0.4$          & $70.7\pm1.8$          & $92.1\pm1.9$          & $88.6\pm1.6$          \\
                                          &                            & IAD                         & $\mathbf{70.2\pm1.4}$ & $\mathbf{75.6\pm1.1}$ & $68.7\pm1.4$          & $\mathbf{95.3\pm0.6}$ & $\mathbf{99.9\pm0.1}$ \\ 
                                          &                            & IAD-Best                    & $71.1\pm1.1$          & $76.9\pm0.5$          & $73.3\pm1.4$          & $96.7\pm0.4$          & $99.9\pm0.0$          \\ \bottomrule 

\end{tabular}
\end{table*}

\begin{figure*}[h]
\centering
\subfloat[Deep SVDD]{\includegraphics[width=0.31\linewidth]{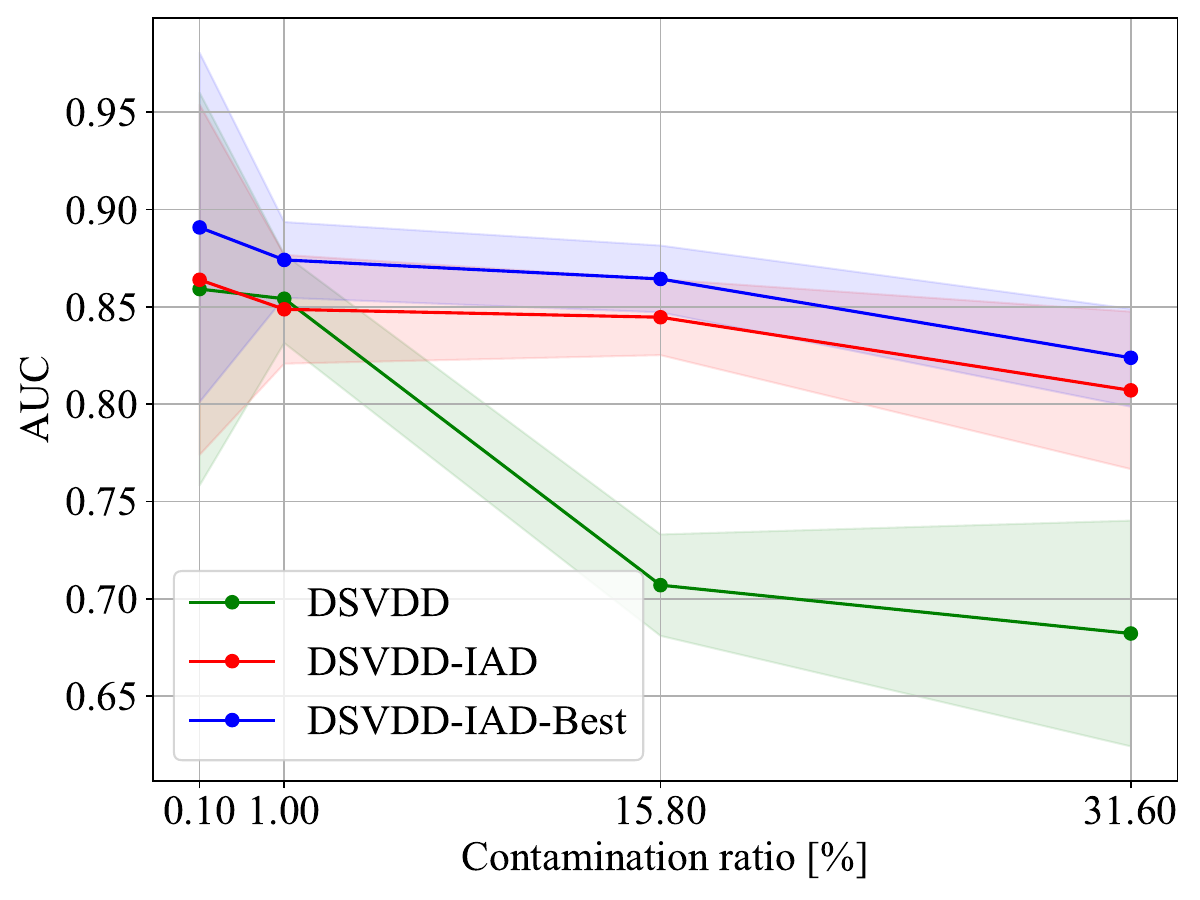}}
\hfil
\subfloat[MAF]{\includegraphics[width=0.31\linewidth]{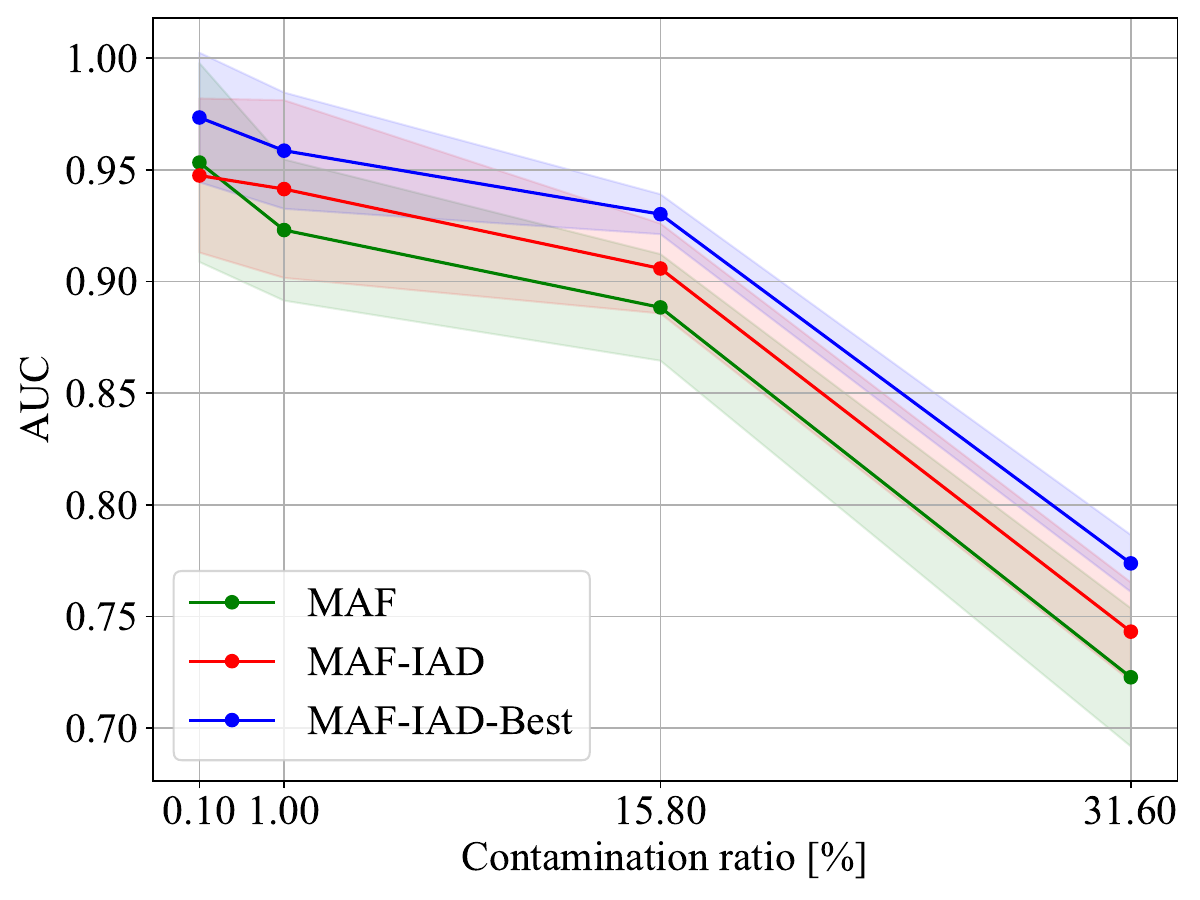}}
\hfil
\subfloat[AE]{\includegraphics[width=0.31\linewidth]{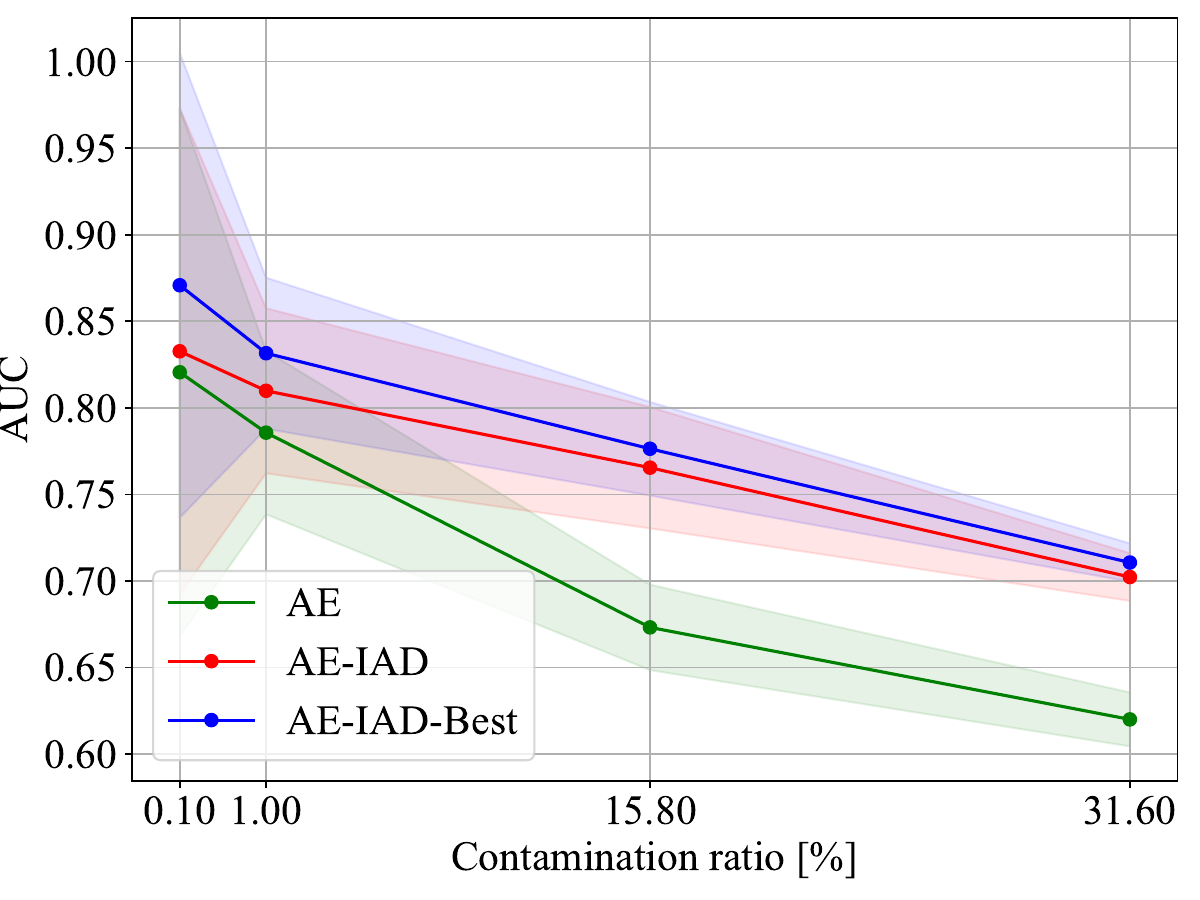}}
\caption{Anomaly detection performance for varying contamination ratios, including 31.6\% (original), 15.8\% (half), 1\%, and 0.1\%}. The colored region represents the standard deviation of the AUC performance. These results are obtained from the Satellite dataset.
\label{fig:satellite-rs}
\end{figure*}

In this section, we extensively validate the robustness of our framework on contaminated datasets with five anomaly detection benchmarks and two image datasets. 
We show the results and analyses of our framework by integrating with 
three representative anomaly detectors, and compare with
the existing robust methods that are proposed to deal with contaminated datasets.

\subsection{Experiment Settings}\label{sec:implementation}

\vspace{2mm}\noindent\textbf{Datasets.}\quad
\emph{Anomaly detection benchmarks}~\cite{rayana2016odds} consist of multivariate tabular datasets. 
As shown in \Tref{tab:dataset}, each of the datasets has $n$ number of data samples with $d$ number of attributes. These benchmarks have a range of contamination ratios from a minimum of $1.2\%$ to a maximum of $31.6\%$. 
We also use the widely used image classification datasets, \emph{MNIST}~\cite{lecun2010mnist} and \emph{Fashion-MNIST} (fMNIST)~\cite{xiao2017/online},
containing approximately 6000 and 6000 training samples per class, respectively, with a total of ten classes. Since \emph{MNIST} and \emph{Fashion-MNIST} are classification datasets, we re-organize the training splits to make anomaly detection scenarios.
In each scenario, we select one class as a normal set and sample all the data in the class set, and add abnormal samples by random sampling from the other nine classes.
In this manner, we generate ten anomaly detection scenarios per image dataset according to classes. 
We use Area Under the Receiver Operating Characteristic Curve (AUC) measured from anomaly scores as an evaluation metric.
For the image datasets, we report the average of per-class AUCs.

\vspace{2mm}\noindent\textbf{Base anomaly detection models of IAD.}\quad
We use the following three methods as a base anomaly detector of IAD: Deep SVDD~\cite{ruff2018deep}, NF~\cite{nachman2020anomaly}, and AE~\cite{sakurada2014anomaly}.
These are 
the main representative approaches in each category:
one-class classification-, probabilistic model-, and reconstruction-based one.
We use \textit{One-Class} Deep SVDD as a base model for the one-class classification method, and \textit{soft-boundary} Deep SVDD incorporating prior knowledge of a contamination ratio is used as a competing method.
For NF, as in the study~\cite{nachman2020anomaly}, the recently proposed masked autoregressive flow (MAF)~\cite{papamakarios2017masked} is used as a base model. 

\begin{table*}[]
\caption{Results on image datasets. 
The contamination ratios are indicated below each dataset name.
We report the average AUC with a standard deviation computed over 5 seeds for each of 10 anomaly detection scenarios.
\textit{Base} denotes
each base anomaly detection model.
IAD represents the model selected by the proposed termination criterion, and 
IAD-Best represents the best model during iterative learning.
The results of IAD with the improvement compared to the base performance are indicated in bold.}\label{tab:table2}
\centering
\begin{tabular}{lllcccccc}
\toprule
\multirow{2}{*}{Approach}                  & \multirow{2}{*}{Base model} & \multirow{2}{*}{Learning method} & \multicolumn{3}{c}{MNIST} &  \multicolumn{3}{c}{fMNIST} \\ \cmidrule(lr){4-9} 
                                           &                             &                                  & 20\%                  & 10\%                  & 5\%                   & 20\%                  & 10\%                  & 5\%                   \\ \midrule
\multirow{3}{*}{One-class classification}  & \multirow{3}{*}{Deep SVDD}  & Base                           & $77.0\pm1.7$          & $82.4\pm1.7$          & $85.8\pm1.9$          & $68.4\pm5.1$          & $73.6\pm4.2$          & $74.6\pm9.1$          \\
                                           &                             & IAD                              & $\mathbf{78.2\pm1.7}$ & $\mathbf{83.1\pm1.0}$ & $85.8\pm1.9$          & $\mathbf{70.6\pm3.5}$ & $\mathbf{74.4\pm2.7}$ & $73.6\pm8.7$          \\
                                           &                             & IAD-Best                         & $80.6\pm1.1$          & $85.8\pm0.9$          & $88.9\pm1.0$          & $75.1\pm1.9$          & $79.0\pm1.4$          & $83.9\pm2.2$          \\ \midrule 

\multirow{3}{*}{Probabilistic model}       & \multirow{3}{*}{MAF}        & Base                           & $90.1\pm2.5$          & $90.7\pm2.2$          & $89.4\pm5.2$          & $87.2\pm0.6$          & $88.7\pm0.5$          & $89.4\pm0.8$  \\
                                           &                             & IAD                              & $\mathbf{91.0\pm2.3}$ & $90.5\pm1.8$          & $\mathbf{89.5\pm5.5}$ & $\mathbf{87.4\pm0.5}$ & $88.5\pm0.6$          & $89.3\pm1.0$  \\
                                           &                             & IAD-Best                         & $91.6\pm1.9$          & $91.0\pm2.2$          & $89.9\pm5.5$          & $88.0\pm0.4$          & $89.3\pm0.5$          & $90.0\pm0.7$  \\ \midrule 

\multirow{3}{*}{Reconstruction}            & \multirow{3}{*}{AE}         & Base                           & $79.6\pm1.8$          & $84.7\pm1.5$          & $87.5\pm1.8$          & $71.9\pm0.8$          & $74.9\pm1.4$          & $79.0\pm1.3$ \\
                                           &                             & IAD                              & $79.6\pm1.5$          & $\mathbf{85.0\pm1.5}$ & $\mathbf{87.7\pm1.6}$ & $71.9\pm1.0$          & $\mathbf{75.1\pm1.4}$ & $78.8\pm1.2$ \\
                                           &                             & IAD-Best                         & $82.0\pm1.0$          & $86.9\pm0.8$          & $89.8\pm1.2$          & $73.4\pm0.8$          & $76.7\pm0.9$          & $80.6\pm1.1$ \\ \bottomrule 
\end{tabular}
\end{table*}

\begin{figure*}[h]
\centering
\subfloat[Deep SVDD]{\includegraphics[width=0.31\linewidth]{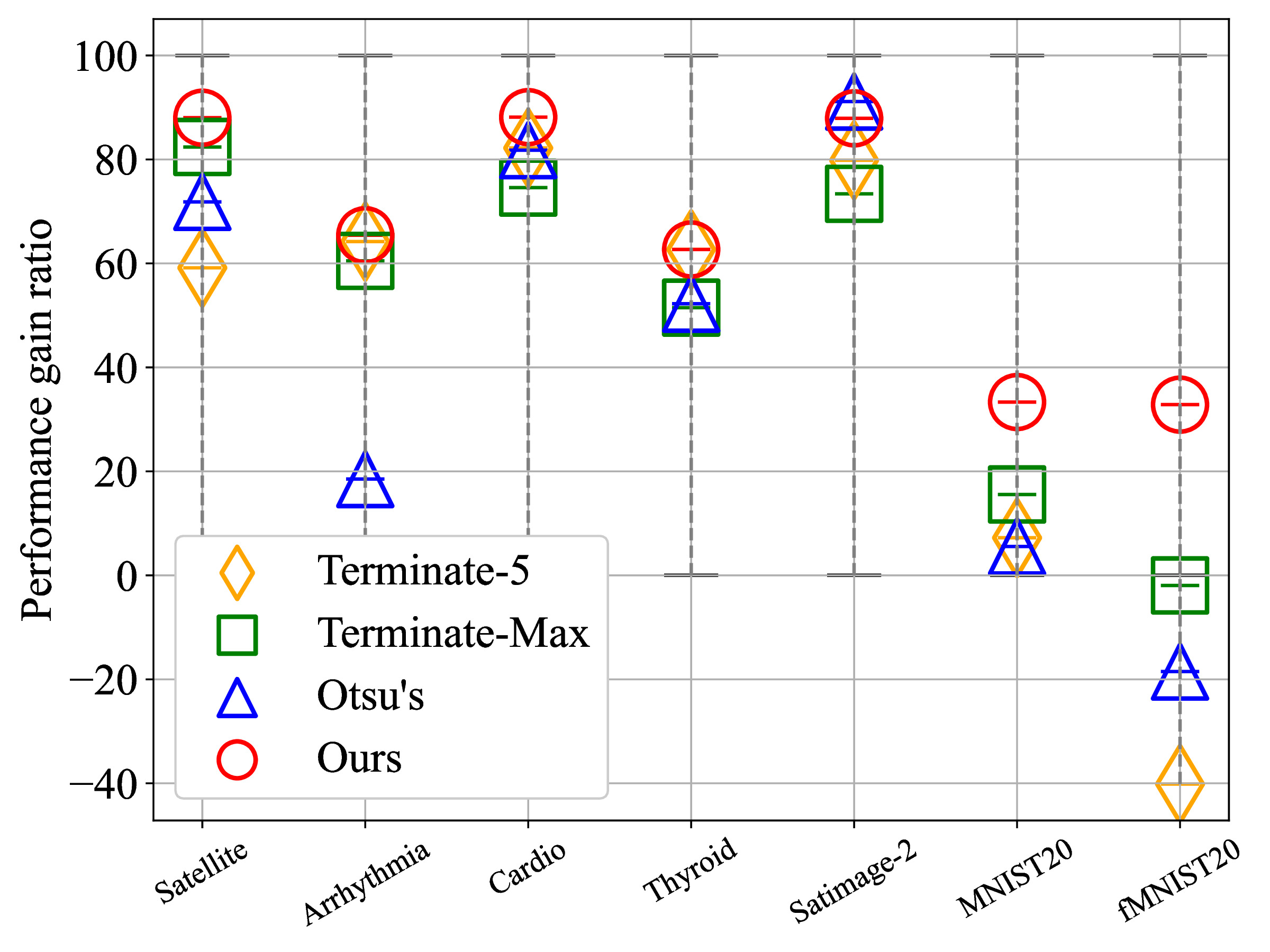}}
\hfil
\subfloat[MAF]{\includegraphics[width=0.31\linewidth]{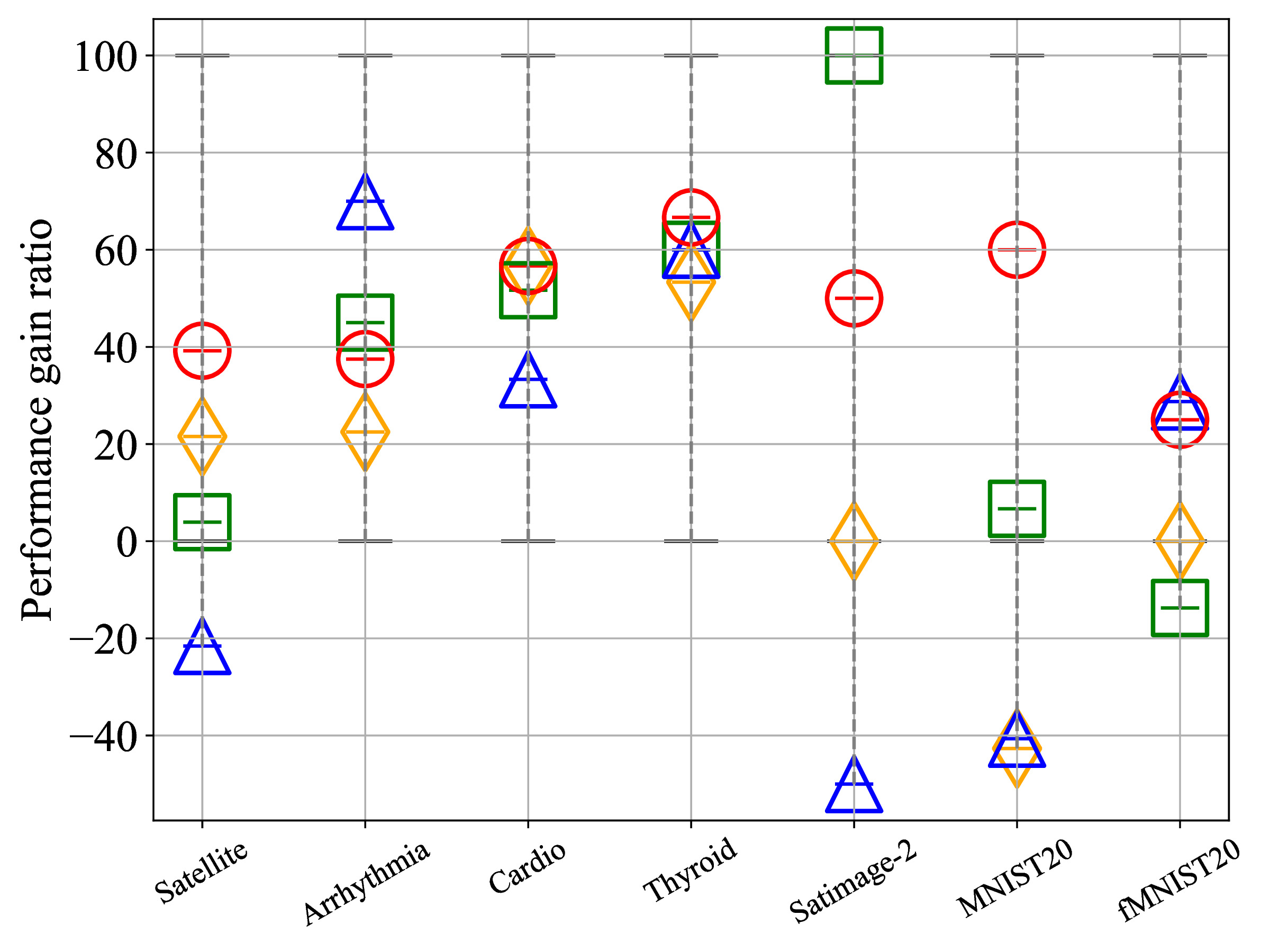}}
\hfil
\subfloat[AE]{\includegraphics[width=0.31\linewidth]{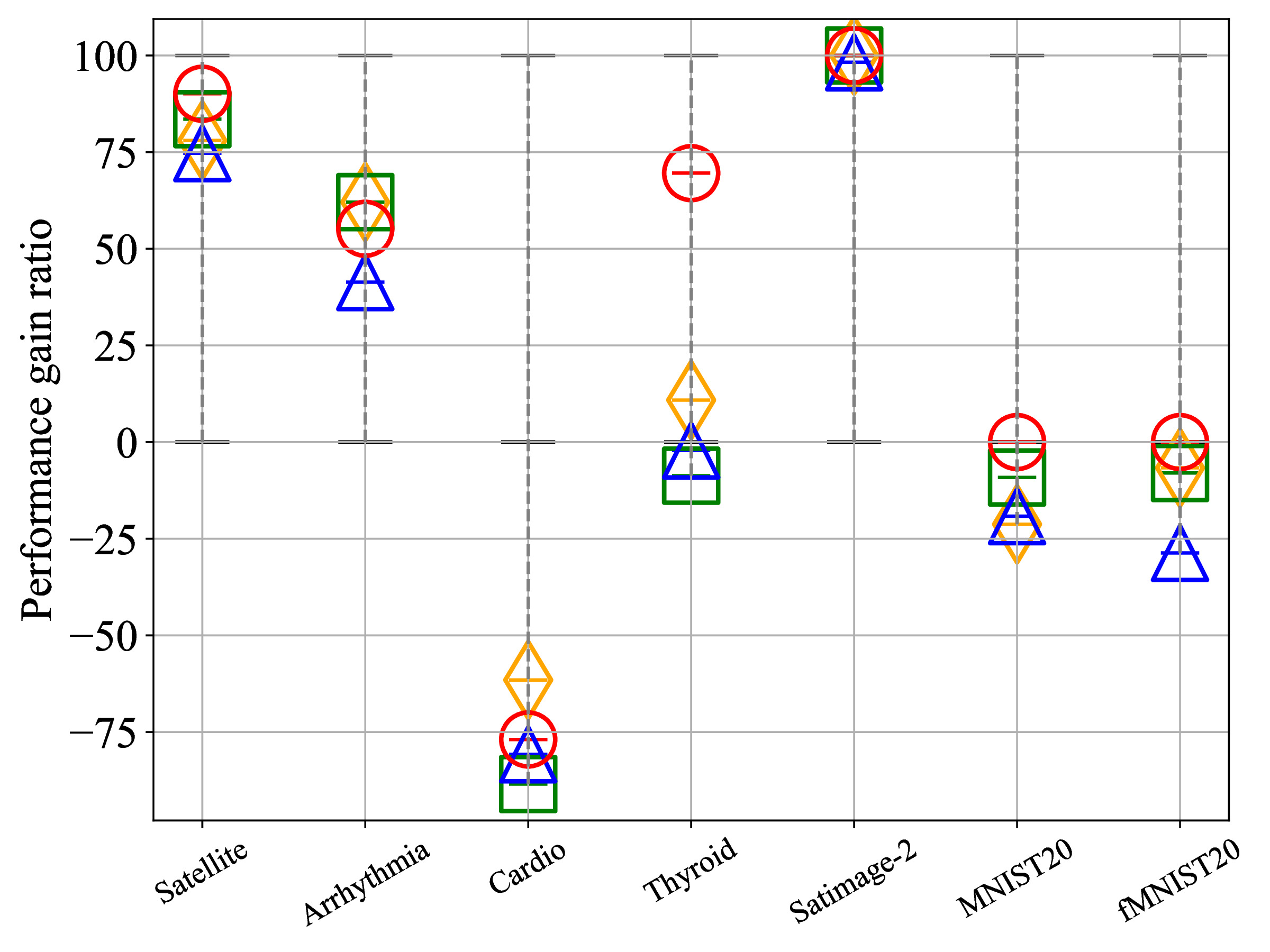}}
\caption{Comparison of termination criteria. 
Each mark in the graphs shows each termination criterion's average performance gain ratio.
The performance gain ratio is a normalized AUC by setting the Base model and IAD-Best performance as 0 and 100 for each dataset.
}
\label{fig:termination}
\end{figure*}

\noindent\textbf{Implementation details and Competing methods.}\quad
To implement the base models of IAD, we use the source code released by the authors and adjust the backbone architectures for each dataset to make a fair comparison. 

For Deep SVDD\footnote{\url{https://github.com/lukasruff/Deep-SVDD-PyTorch}} and AE, we use standard MLP feed-forward architectures for the tabular datasets and
LeNet~\cite{lecun1989backpropagation}-type convolutional neural networks (CNNs) for the image datasets following Ruff~\etal\cite{ruff2019deep}.
For Deep SVDD, a 3-layer MLP with 32-16-8 units is used on the \textit{Satellite}, \textit{Cardio}, and \textit{Satimage-2} dataset, a 3-layer MLP with 128-64-32 units is used on the \textit{Arrhythmia} dataset, and a 3-layer MLP with 32-16-4 units is used on the \textit{Thyroid} dataset.
On \textit{MNIST}, a CNN with two modules, $8\times(5\times5)$-filters followed by $4\times(5\times5)$-filters, and a final dense layer of 32 units are used.
On \textit{Fashion-MNIST}, $16\times(5\times5)$-filters followed by $32\times(5\times5)$-filters and two dense layers of 128 and 64 units are used. 
Each convolutional module is followed by Leaky ReLu activation and $(2\times2)$-max-pooling.
For AE, we utilize the above architectures for an encoder network and implement a decoder network symmetrically. In a decoder, the max-pooling and convolutions are replaced with simple upsampling and deconvolutions.

For MAF\footnote{\url{https://github.com/kamenbliznashki/normalizing\_flows}},
a network consisting of five autoregressive layers with 32 hidden units is used with a standard Gaussian as a base density distribution, for all the datasets.
In the case of image data, data is flattened into a vector form for input. 
We use Adam optimizer~\cite{kingma2014adam} with a batch size of $128$ and $200$ with a learning rate of $10^{-3}$ and $10^{-4}$ in the 
benchmarks and image datasets, respectively.
We set the maximum number of rounds $T$ as 15 and 10 for the tabular and image datasets, respectively. 
The temperature parameter $1/\tau$ is set to 4.
For the 
benchmarks, samples are standardized to have zero mean and unit variance. 
For the image datasets, each image data is normalized using the $L_1$-norm and rescaled to $[0, 1]$ via min-max scaling.

Our framework, IAD, is compared with the following six unsupervised methods: \textit{soft-boundary} (SB) Deep SVDD~\cite{ruff2018deep}, OneFlow\footnote{\url{https://github.com/gmum/OneFlow}}~\cite{maziarka2021oneflow}, DRAE~\cite{xia2015learning},
RSRAE\footnote{\url{https://github.com/dmzou/RSRAE}}~\cite{lai2020robust},
SDOR~\cite{pang2020self}, and DAGPR\footnote{\url{https://github.com/fanjinan/DAGPR}}~\cite{fan2020robust}.
The competing methods include the iterative learning methods with pseudo-labeled data~\cite{xia2015learning, pang2020self, fan2020robust}, the method using a prior-knowledge as a hyper-parameter~\cite{ruff2018deep, maziarka2021oneflow}, and the method based on a modified architecture~\cite{lai2020robust}.
We use a true contamination ratio as the parameter $\nu$ of SB Deep SVDD and $\alpha$ of OneFlow, and for the other methods, the suggested optimal parameters are used.  
We utilize the authors' source code for SB Deep SVDD, OneFlow, RSRAE, and DAGPR.
For DRAE, we reproduce the model following the details of the authors' paper using the same AE architecture described in this subsection.
For SDOR, the same AE architecture is used for the initial pseudo-labeling, and a binary regression model is trained iteratively using pseudo-normal/abnormal data selected in the manner suggested by the authors. 
All the experiments are conducted on a machine with Intel Xeon Silver 4210 CPU with GeForce GTX 1080Ti GPU.

\subsection{Experiment results of IAD}
In this subsection, we analyze the effectiveness of IAD
on the anomaly detection benchmarks (\Tref{tab:table1}) and image datasets (\Tref{tab:table2}).
\textit{IAD} refers to our method with 
the proposed termination criterion, and \textit{IAD-Best} refers to the model with the highest performance across all training rounds.

\begin{table*}[t]
\caption{Results on datasets with an anomaly ratio of 0.1\%.
We report the average AUC with a standard deviation computed over 10 seeds for anomaly detection benchmarks and 5 seeds for image datasets.
\textit{Base} denotes each base anomaly detection model.
IAD represents the model selected by the proposed termination criterion,
and IAD-Best represents the best model during iterative learning.
The results of IAD with the improvement compared to the base performance are shown in bold.}\label{tab:table0.1}
\centering
\begin{tabular}{llccccccc}
\toprule
\multirow{2}{*}{Base model} & \multirow{2}{*}{Learning method}  & Satellite             & Arrhythmia            & Cardio                & Thyroid               & Satimage-2    & MNIST         & fMNIST        \\ \cmidrule(lr){3-9} 
                            &                                   & \multicolumn{7}{c}{0.1\%}                                                                                                                     \\ \midrule
\multirow{3}{*}{Deep SVDD}  & Base                              & $85.9\pm10.1$         & $78.7\pm6.3$          & $77.5\pm16.9$         & $79.3\pm12.4$         & $95.4\pm6.4$          & $91.0\pm6.1$          & $80.2\pm7.9$  \\
                            & IAD                               & $\mathbf{86.4\pm9.0}$ & $\mathbf{81.1\pm10.3}$& $\mathbf{82.4\pm23.3}$& $\mathbf{82.1\pm9.2}$ & $93.9\pm5.9$          & $\mathbf{91.2\pm4.3}$ & $78.1\pm12.3$ \\ 
                            & IAD-Best                          & $89.1\pm9.0$          & $90.9\pm8.2$          & $95.1\pm8.6$          & $93.3\pm4.6$          & $98.4\pm3.7$          & $96.4\pm2.3$          & $89.7\pm5.6$  \\ \midrule  

\multirow{3}{*}{MAF}        & Base                              & $95.3\pm4.5$          & $99.6\pm0.2$          & $87.8\pm15.5$         & $97.6\pm2.4$          & $99.3\pm0.7$          & $85.5\pm7.9$          & $90.5\pm7.6$  \\
                            & IAD                               & $94.7\pm3.4$          & $99.6\pm0.1$          & $\mathbf{92.8\pm9.5}$ & $97.5\pm2.0$          & $\mathbf{99.4\pm0.4}$ & $\mathbf{86.4\pm6.7}$ & $\mathbf{91.0\pm6.5}$  \\  
                            & IAD-Best                          & $97.3\pm2.9$          & $99.8\pm0.2$          & $94.8\pm8.4$          & $98.5\pm1.5$          & $99.8\pm0.2$          & $87.4\pm6.7$          & $93.5\pm4.2$  \\ \midrule 

\multirow{3}{*}{AE}         & Base                              & $82.1\pm15.3$         & $82.2\pm19.3$         & $84.6\pm24.9$         & $96.7\pm5.0$          & $97.6\pm2.3$          & $92.9\pm4.0$          & $86.8\pm5.3$  \\
                            & IAD                               & $\mathbf{83.3\pm14.0}$& $82.2\pm19$           & $\mathbf{88.3\pm21.6}$& $94.0\pm4.8$          & $\mathbf{99.8\pm0.3}$ & $\mathbf{93.1\pm4.8}$ & $86.2\pm5.8$  \\ 
                            & IAD-Best                          & $87.1\pm13.4$         & $86.4\pm18.6$         & $88.7\pm22.1$         & $97.2\pm3.7$          & $99.9\pm0.2$          & $96.3\pm2.3$          & $88.4\pm5.1$  \\ \bottomrule 
\end{tabular}
\end{table*}

\begin{table*}[t]
\caption{
Comparison between the ensemble model and IAD.
The contamination ratio is indicated below each dataset name.
We report the average AUC with a standard deviation computed over 10 seeds.
\textit{Base} represents AE in this table. \textit{Ensemble}+IAD represents the ensemble model selected by the proposed termination criterion during iterative learning.
The performance of IAD is presented for a reference purpose.
The highest performances among \textit{Base}, \textit{Ensemble}, and \textit{Ensemble}+IAD
are indicated in bold. 
IAD performances higher than \textit{Ensemble} are also indicated in bold.
}\label{tab:table-ensem}
\centering
\begin{tabular}{lccccccc}
\toprule
\multirow{2}{*}{Learning method}    & Satellite & Arrhythmia & Cardio & Thyroid & Satimage-2 & MNIST & fMNIST   \\ \cmidrule(lr){2-8} 
                                    & 31.6\%    & 14.6\%     & 9.6\%  & 2.5\%   & 1.2\%      & 20\%  & 20\%     \\ \midrule
Base                                & $62.0\pm1.6$          & $74.0\pm0.4$          & $70.7\pm1.8$          & $92.1\pm1.9$          & $88.6\pm1.6$          & $79.6\pm1.8$          & $\mathbf{71.9\pm0.8}$ \\
Ensemble                            & $63.1\pm0.9$          & $74.2\pm0.2$          & $\mathbf{71.4\pm1.1}$ & $\mathbf{94.8\pm0.9}$ & $89.9\pm1.2$          & $79.9\pm0.8$          & $71.7\pm0.6$          \\ 
Ensemble+IAD                        & $\mathbf{69.5\pm0.7}$ & $\mathbf{76.1\pm0.4}$ & $69.6\pm1.2$          & $93.8\pm1.3$          & $\mathbf{99.9\pm0.0}$ & $\mathbf{80.1\pm0.9}$ & $\mathbf{71.9\pm0.6}$ \\ \midrule 
IAD                                 & $\mathbf{70.2\pm1.4}$ & $\mathbf{75.6\pm1.1}$ & $68.7\pm1.4$          & $\mathbf{95.3\pm0.6}$ & $\mathbf{99.9\pm0.1}$ & $79.6\pm1.5$          & $\mathbf{71.9\pm1.0}$ \\ 
\bottomrule                

\end{tabular}
\end{table*}

\begin{figure*}[t]
\centering
\subfloat[Changes in importance weights]
{\includegraphics[width=0.25\linewidth]{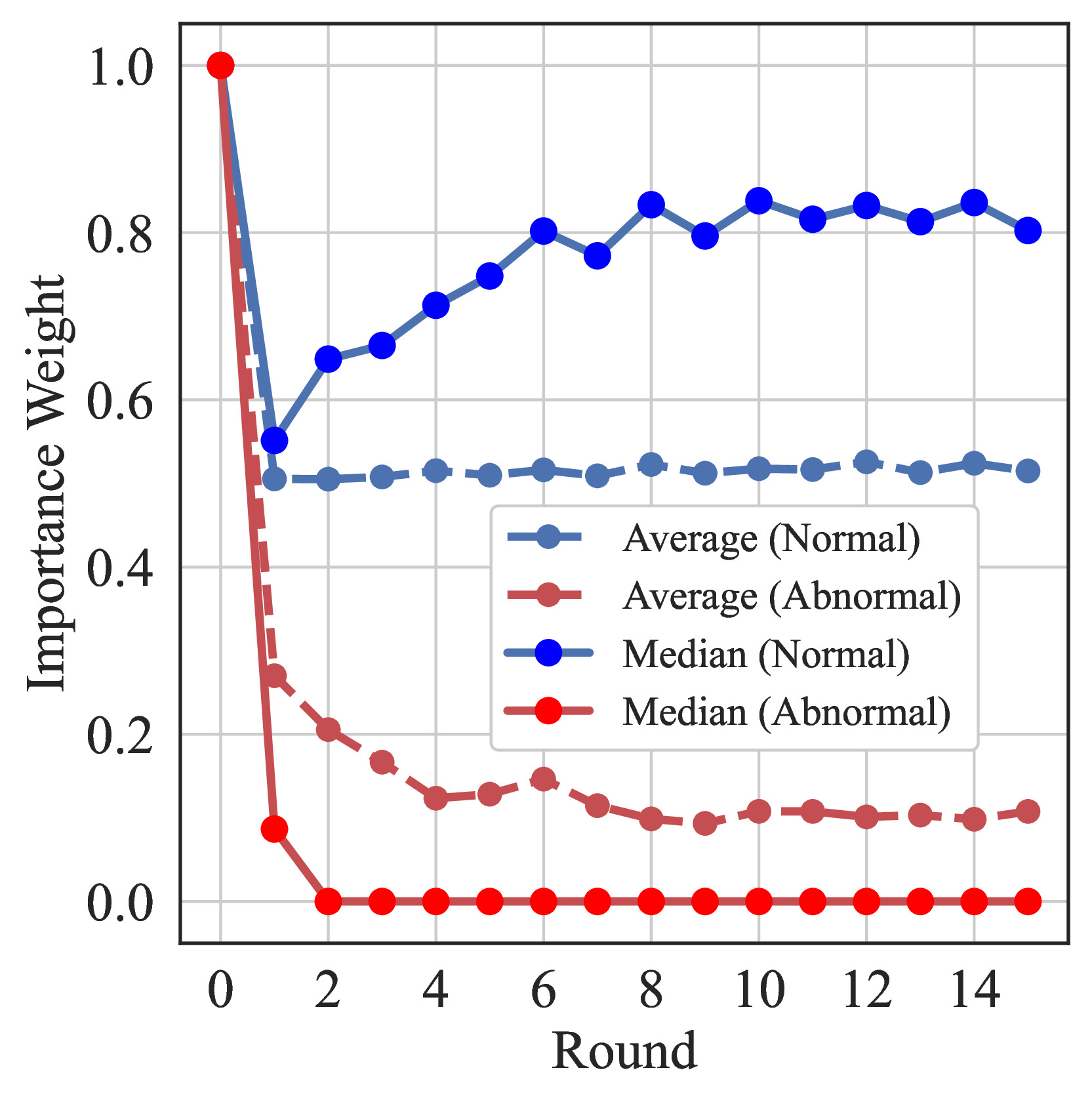}\label{fig:psl_confi}}
\hfil
\subfloat[Changes in the distribution of importance weights]
{\includegraphics[width=0.25\linewidth]{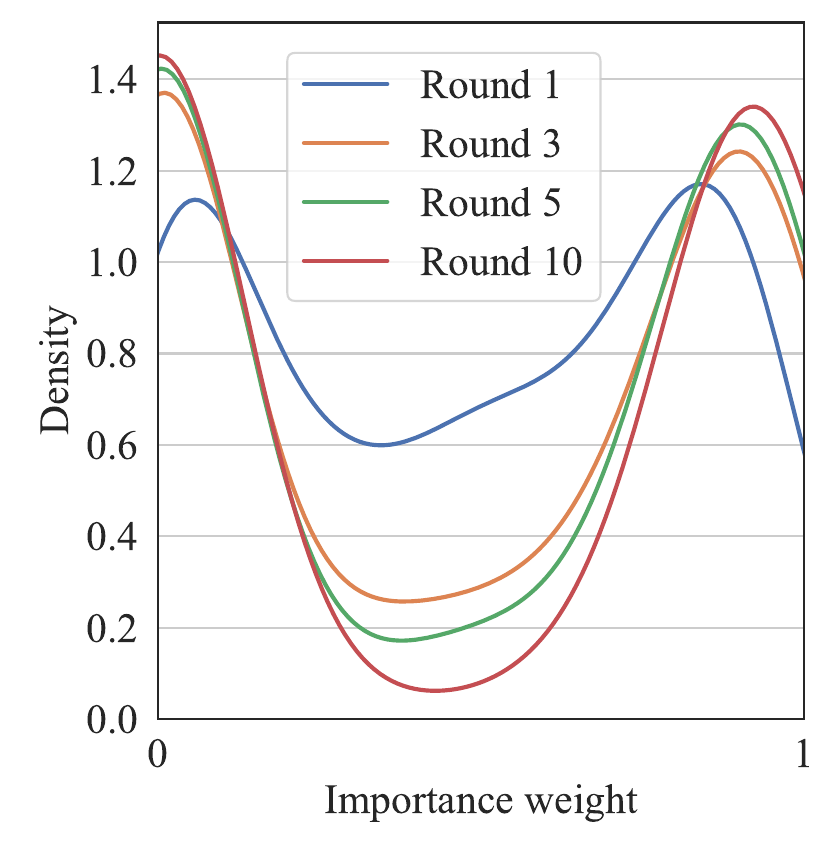}\label{fig:self-enforcing}}
\hfil
\subfloat[Comparison of t-SNE distribution]{\includegraphics[width=0.44\linewidth, height=5.1cm]{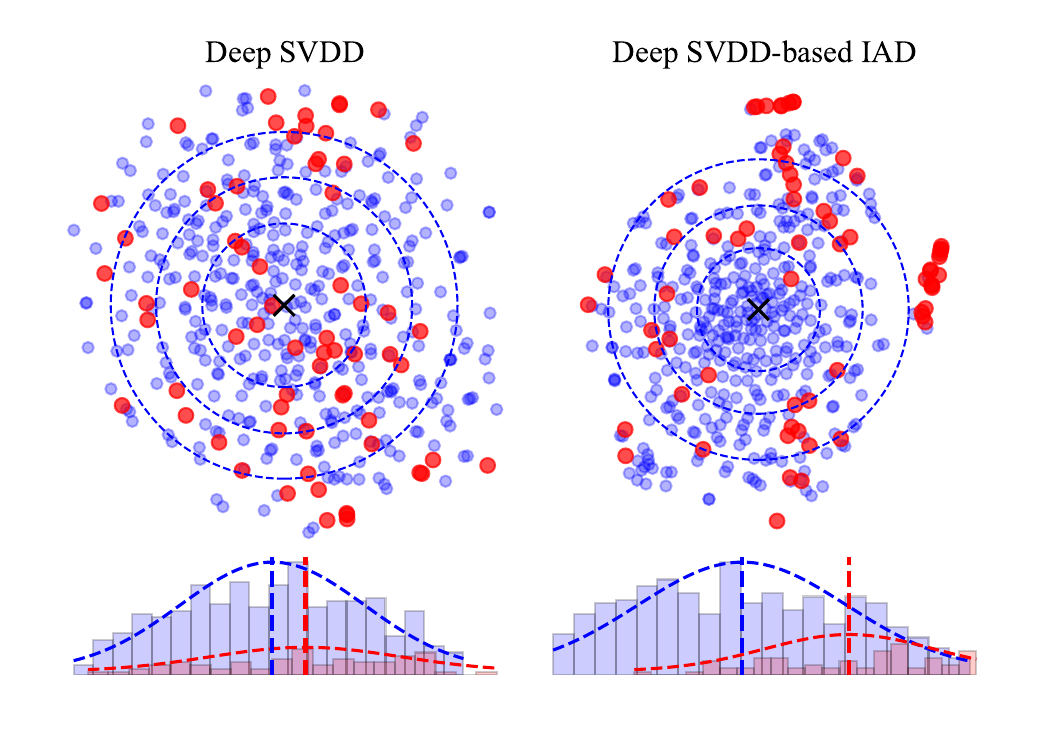}\label{fig:z_hidden}}
\caption{\textbf{(a)} Changes in importance weights of IAD with Deep SVDD on Arrhythmia. Here, the average and median importance weight of true normal and true abnormal data over rounds are shown.
\textbf{(b)} Changes in the distribution of importance weights of the same model with (a). The distribution is estimated by KDE.
\textbf{(c)} t-SNE distributions on the top and the histograms of the anomaly score on the bottom obtained from base Deep SVDD and IAD. The concentric circles represent the boundary containing $25\%$, $50\%$, and $75\%$ of the total data from the center of the hypersphere.
These results are obtained from Arrhythmia.}
\label{fig:vis}
\end{figure*}

\begin{figure*}[t]
\centering
\subfloat[Deep SVDD]{\includegraphics[width=0.33\linewidth]{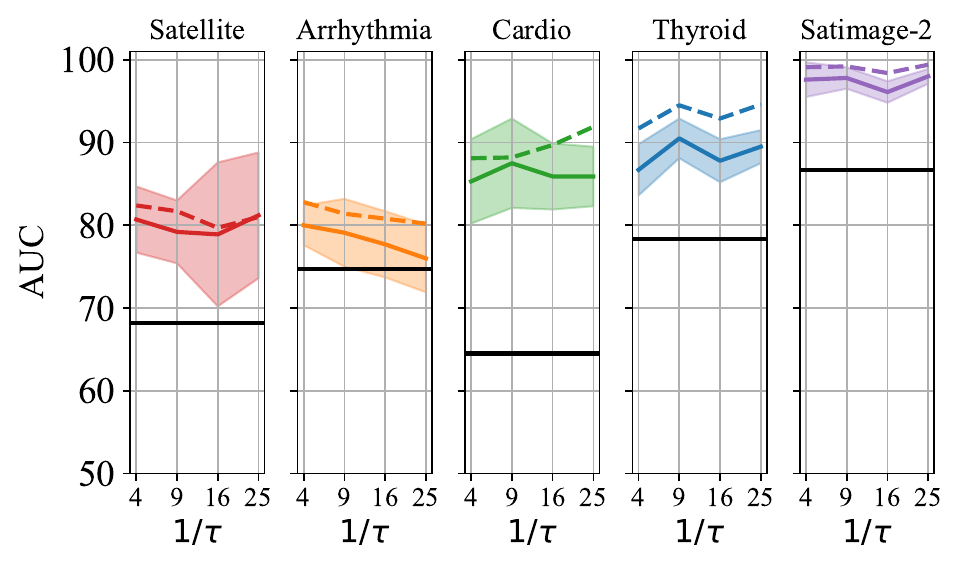}}
\hfil
\subfloat[MAF]{\includegraphics[width=0.33\linewidth]{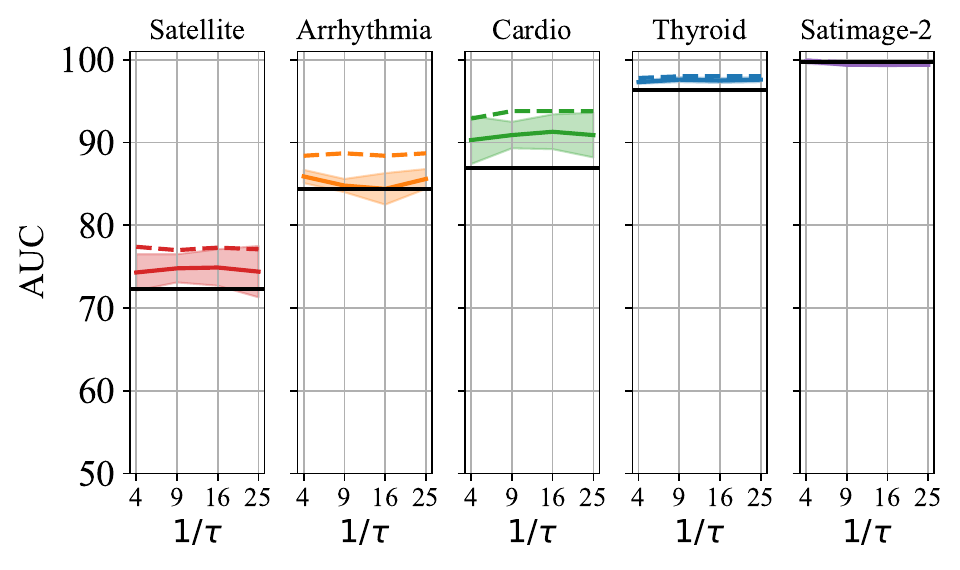}}
\hfil
\subfloat[AE]{\includegraphics[width=0.33\linewidth]{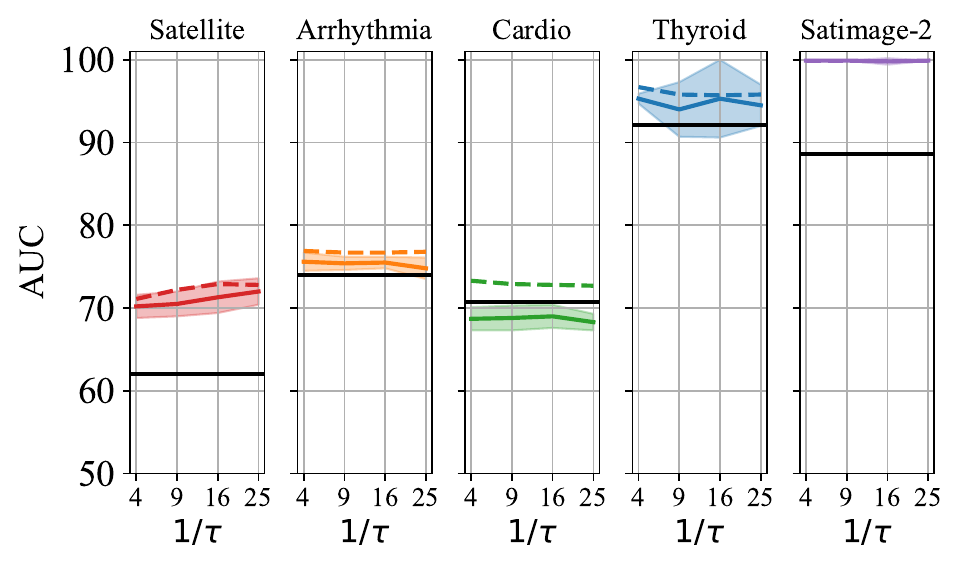}}
\caption{
Performance according to $\tau$ on the anomaly detection benchmarks.
The black lines, solid color lines, and dashed color lines represent the average AUC of base anomaly detection models, IAD, and IAD-Best, respectively.
The shaded areas represent the standard deviation of the AUC of IAD.}
\label{fig:alpha}
\end{figure*}

\subsubsection{Effect of contaminated datasets}
We empirically investigate the effect of contaminated datasets on anomaly detection performance (Tables~\ref{tab:table1} and \ref{tab:table2}).
The results show the common expectation that, as the contamination ratio increases, the performance tends to be lower.

However, to clarify the observation of the trend in the tabular datasets,
we additionally evaluate performances by adjusting the contamination ratio within one dataset. For this experiment, we use the \textit{Satellite} dataset, which has the highest contamination ratio. 
The performance-contamination ratio trend is shown for all three base models (\Fref{fig:satellite-rs}).
It indicates that the base models designed with the normality assumption are adversely affected by contaminated datasets.
On the other hand, IAD alleviates the performance degradation caused by contamination.

\subsubsection{Performance of IAD}
In most cases, IAD improves performance and reduces performance variance of the base models (\Tref{tab:table1} and \Tref{tab:table2}).
The biggest performance improvement is obtained with Deep SVDD, followed by AE and MAF. 
It is notable that IAD-Best performance is higher than the Base models in all the scenarios. 
It validates that the proposed iterative learning improves the robustness of the base models when a dataset is contaminated. However, because of contamination, the performance fluctuates during the iterative training. Therefore, when to terminate the iterative training is a non-trivial problem. A termination rule is a significant part of the iterative learning framework to capture this improvement. 

\subsubsection{Termination criterion}\label{sec:termination}
The proposed termination criterion obtains on average 58.7\% of the performance improvement compared to the maximum performance improvement for all the datasets.
We compare the proposed termination criterion and the termination criteria used by the existing studies~\cite{pang2020self, xia2015learning} dealing with contaminated datasets (\Fref{fig:termination}). 
The competing criteria include the termination method that terminates at a fixed round (fifth round~\cite{pang2020self} or the last round) and the termination method based on Otsu's method~\cite{otsu1979threshold, xia2015learning}. 
We denote each method as Terminate-5, Terminate-Max, and Otsu's.
The termination criteria are compared by performance gain ratio (PGR) which is a normalized AUC by setting the Base model and IAD-Best performance as 0 and 100: 
\begin{equation*}
    \textrm{PGR} = 100\times\frac{(\textrm{AUC(IAD)} - \textrm{AUC(Base)})}{(\textrm{AUC(IAD-Best)} - \textrm{AUC(Base)})}.
\end{equation*}
As shown in \Fref{fig:termination}, the proposed termination criterion obtains the highest performance gain in most cases. 
Terminate-5, Terminate-Max, and Otsu's methods obtain the performance gains of 46.1\%, 50.1\%, and 35.9\% on average.

However, for AE, there are few cases in which the terminated performance is similar to or lower than that of a Base model.
In these cases, the IAD-Best performance is obtained at the early stage of the iterative learning, and the performance decreases thereafter. 
We conjecture this is due to a limitation of the base anomaly detector. It is shown in the recent study~\cite{yoon2021autoencoding} that AE does not model the likelihood of data. Thereby, it can also reconstruct unseen data resulting in failure cases in anomaly detection when regularization is not imposed.

\subsubsection{Performance of IAD on a very low anomaly ratio}
Although the normality assumption is difficult to hold in most real-world scenarios, some cases with very low anomaly ratios approximate the normality assumption.
To validate our framework under scenarios with the normality assumption,
we evaluate our framework on datasets with an extremely low contamination ratio of 0.1\% (\Tref{tab:table0.1}).
The randomly sampled anomalies constitute 0.1\% of each dataset.
For the \textit{Arrhythmia} dataset, only one abnormal sample is included.
Although the performance gain ratio is smaller than the ones from the previous experiments,
IAD still improves the performance of the base model in the majority of cases.
However, when a low contamination ratio and a relatively high performance of the base model, the value of $h$ calculated based on the halfway point may not vary significantly regardless of the improvement in model performance.
Therefore, the model with a small performance gain as well as the model that has not gone through enough rounds may be selected.
The large performance variance is caused by the randomness in constructing extremely imbalanced datasets, \ie very few anomalies are randomly sampled.

\subsubsection{Comparison with an ensemble model}
One commonly used approach to deal with noise in a dataset is an ensemble method.
Therefore, we additionally apply an ensemble model to our framework,
which is known to achieve a better generalization capability than that of a single model~\cite{dietterich2000ensemble}.  
In this sense, the anomaly score obtained from an ensemble model is expected to be robust to a contaminated dataset.
We instantiate the ensemble model by five AEs for this experiment. Each AE is trained on randomly sampled sub-datasets covering 80\% of the given dataset.\footnote{For ensemble experiments, we use partial datasets of the entire datasets used for IAD. Thus, direct comparison between IAD and Ensemble + IAD is unfair.} Anomaly scores are scaled to have the median value of 1 and then averaged for the aggregation.
When applying the ensemble model to IAD, ensemble-based training and anomaly score computation through ensemble aggregation are performed in lines 4-6 and 7 of \Aref{alg:iad}. Each AE is trained with updated importance weights corresponding to a sub-dataset. The results are shown in \Tref{tab:table-ensem}.

We address three questions in this experiment: the compatibility of IAD with the ensemble model, the comparison between IAD and the ensemble model, and the comparison between IAD and ensemble model-based IAD. 
The ensemble model improves the performance of the base model in most cases. 
When IAD is applied to the ensemble model, the performance further improves in 5 out of 7 datasets which validates the compatibility of IAD with ensemble models.
In addition, IAD without an ensemble approach performs favorably against the ensemble model in most cases (bold numbers in \Tref{tab:table-ensem}).
However, there is no substantial performance gap between IAD with and without the ensemble approach. 
When an ensemble model is trained with IAD, the individual models in an ensemble are already robust to the data noise. Because of this robustness, a further aggregation of individual model outputs does not give a meaningful difference.
Although an ensemble approach does not improve the performance of IAD, one notable advantage of using an ensemble approach is lower performance variance.

\begin{table*}[t]
\caption{Performances of the competing methods. 
The contamination ratio is indicated below each dataset name.
On anomaly detection benchmarks and image datasets, the average AUC is computed over 10 and 5 seeds, respectively.
For image datasets, performance is evaluated over 10 anomaly detection scenarios. 
For each dataset, the highest performance is indicated in bold.
\label{tab:table3}}
\centering
\begin{tabular}{lccccccc}
\toprule

\multirow{2}{*}{Method}     & Satellite & Arrhythmia & Cardio & Thyroid & Satimage-2 & MNIST & fMNIST \\ \cmidrule(lr){2-8}
                            & 31.6\%    & 14.6\%     & 9.6\%  & 2.5\%   & 1.2\%      & 20\%  & 20\%   \\ \midrule

SB Deep SVDD~\cite{ruff2018deep}        & $62.4\pm6.7$              & $62.5\pm2.3$          & $62.3\pm9.4$          & $81.3\pm3.7$          & $81.9\pm5.5$          & $74.6\pm1.9$          & $59.1\pm11.8$         \\
OneFlow~\cite{maziarka2021oneflow} & $75.3\pm1.0$              & $75.9\pm0.8$          & $76.8\pm4.6$          & $85.4\pm4.8$          & $88.3\pm4.8$          & $83.4\pm1.1$          & $77.0\pm0.9$          \\
DRAE~\cite{xia2015learning}             & $74.1\pm1.1$              & $77.2\pm1.9$          & $86.6\pm6.8$          & $65.8\pm8.6$          & $99.5\pm0.5$          & $78.2\pm3.2$          & $78.9\pm1.5$          \\    
RSRAE~\cite{lai2020robust}              & $67.5\pm0.6$              & $75.4\pm0.5$          & $69.3\pm1.2$          & $90.5\pm3.1$          & $88.8\pm2.0$          & $82.6\pm1.6$          & $70.8\pm0.9$          \\
SDOR~\cite{pang2020self}                & $62.1\pm1.2$              & $74.1\pm0.5$          & $70.6\pm1.7$          & $92.2\pm1.7$          & $88.7\pm2.4$          & $80.4\pm9.5$          & $75.1\pm1.0$          \\ 
DAGPR~\cite{fan2020robust}              & $66.0\pm5.9$              & $80.7\pm2.5$          & $86.6\pm4.3$          & $71.0\pm8.0$          & $87.2\pm2.5$          & $77.4\pm3.6$          & $85.4\pm2.9$          \\ 
\hline

\textbf{IAD (Deep SVDD)}                & $\mathbf{80.7\pm4.0}$     & $80.0\pm2.4$          & $85.3\pm5.1$          & $86.7\pm3.1$          & $97.6\pm2.1$          & $78.2\pm1.7$          & $70.6\pm3.5$          \\ 
\textbf{IAD (AE)}                       & $70.2\pm1.4$              & $75.6\pm1.1$          & $68.7\pm1.4$          & $95.3\pm0.6$          & $\mathbf{99.9\pm0.1}$ & $79.6\pm1.5$          & $71.9\pm1.0$          \\ 
\textbf{IAD (MAF)}                      & $74.3\pm2.2$              & $\mathbf{85.9\pm0.8}$ & $\mathbf{90.3\pm2.9}$ & $\mathbf{97.3\pm0.2}$ & $99.8\pm0.3$          & $\mathbf{91.0\pm2.3}$ & $\mathbf{87.4\pm0.5}$ \\ \bottomrule

\end{tabular}
\end{table*}

\subsection{Analysis}
\subsubsection{Importance weight} 
We show the importance weights of the contaminated dataset, being updated in the iterative learning process (\Fref{fig:psl_confi}).
In this graph, we divide the training data as true normal (blue) and true abnormal (red) using labels to visualize the change in the importance weights in IAD.
The initial value of the importance weights of all the data is set to 1. 
The average and median plots of the importance weights for each data group are clearly separable with a substantial gap.
From this observation, we hypothesize a self-enforcing effect: 
Samples with low weights should get even lower weights in the next iteration since the networks optimize less for these samples leading to higher anomaly scores. 
The effect is also reversed for samples with high weights.
To confirm the effect, we investigate the distribution of importance weights in each round (\Fref{fig:self-enforcing}). 
The distribution is estimated using KDE~\cite{parzen1962estimation}.
The estimated distribution of importance weights is gradually skewed left and right over rounds, showing a sharp bimodal shape.
This analysis implies
that importance weights correctly enforce the learning signals from normal samples while suppressing the signals from anomalous samples in our framework.

\subsubsection{Feature distribution} 
We visualize the latent feature space mapped to two-dimensional space via t-SNE~\cite{van2008visualizing} (\Fref{fig:z_hidden}).
Here, for ease of interpretation of feature distribution, we choose Deep SVDD as a base anomaly detector of IAD.
The feature distribution of the base model, \ie, Deep SVDD (\Fref{fig:z_hidden} left side), and the feature distribution of IAD (\Fref{fig:z_hidden} right side) are shown.
The difference is clear between the feature distributions of the Base model and IAD. 
The normal samples are densely distributed around the center for IAD, while the features are relatively scattered for the Base model.
For a clearer interpretation of the t-SNE distribution, we plot a histogram of anomaly scores, representing each sample's distance from the hypersphere center of Deep SVDD.
For reference, we fit a Gaussian distribution to each histogram and visualize its center.
By applying our framework, the distribution of normal and abnormal samples are distinctly separated,
leading to performance improvement.

\subsubsection{Temperature parameter $\tau$}
We investigate the effect of $\tau>0$ (\Eref{eqn:ab}), which affects the distribution of importance weights.
Note that the temperature $\tau$ is an inevitable hyper-parameter when using a softmax or sigmoid function.
We show the performance of the proposed framework according to $1/\tau\in\{4, 9, 16, 25\}$
(\Fref{fig:alpha}).
Although the performance fluctuates in some cases with Deep SVDD, 
the performance of IAD (solid color lines) and IAD-Best (dashed color lines) are insensitive to $\tau$ in most of the cases, along with improved performance compared to the base model (black lines).

\subsection{Comparison with competing methods}
In \Tref{tab:table3}, we compare AUC performance of our method with the competing robust methods
mentioned in Sec.~\ref{sec:implementation}. 
The models trained with our framework IAD show the highest performance for all the datasets.
The competing methods are categorized according to the underlying approach as described in \ref{sec:related works-b}, which are one-class classification-, reconstruction-, regression-, and probabilistic model-based ones.

First, the performance of \textit{One-Class} deep SVDD-based IAD is notably higher than that of \textit{soft-boundary} deep SVDD, 
although the true contamination ratio was used for the hyper-parameter of \textit{soft-boundary} deep SVDD.
In most cases, IAD also performed favorably against OneFlow, which uses the true contamination ratio.
This comparison shows that even with a good hyper-parameter setting, the proposed learning framework is 
more robust than the competitors.

Second, among the reconstruction-based methods, none of them performs best for all the cases. 
AE-based IAD does not stand out in comparison with DRAE in terms of performance, but when compared to RSRAE, it obtained higher performance in the majority of datasets.
In our case, we do not modify architecture or use pseudo-labels but only rely on the learning mechanism to improve the na\"ive AE. Therefore, the limited performance of AE-based IAD may come from AE's lack of ability to explicitly predict the likelihood of data~\cite{yoon2021autoencoding}.

We also report the regression-based robust methods, DAGPR~\cite{fan2020robust} and SDOR~\cite{pang2020self}, for reference.
The regression-based methods use binary class labels, \ie, labels from both normal and abnormal samples. In a contaminated setting, the requirement of binary labels introduces hyper-parameters to control the amounts of pseudo-labels, which are sensitive to data characteristics. 
Since one of our goals is to avoid such hyper-parameter selection, we do not apply our framework to the regression-based methods.
All our IAD variants perform favorably against the regression-based methods.

Lastly, MAF-based
IAD shows the highest performance among all the competing methods in most cases.
Considering that the majority of anomaly detection studies are based on AE-based methods, our study empirically shows that NF-based methods can be more 
robust to contaminated datasets, especially when combined with our iterative method.

\section{Conclusion}\label{sec:conclusion}
In this work, we present an unsupervised anomaly detection framework targeted for contaminated data scenarios. 
An iterative learning method of a base deep anomaly detection model is proposed to improve the robustness against a contaminated dataset. We introduce importance weights in the iterative learning process to alleviate the negative learning effects of anomalous samples. 
In addition, we propose a new termination criterion based on the training dynamics of anomaly scores to capture a robust representation of a model during iterative training.
The effectiveness of the proposed framework, including the termination criterion, is validated with extensive experiments on five anomaly detection benchmarks and two image datasets. 
Our framework shows robust performance under a wide 
range of contamination ratios without any requirements for sensitive hyper-parameter tuning or prior knowledge of contamination ratios.

\section*{Acknowledgments}
This work was supported by Institute of Information \& communications Technology Planning \& Evaluation (IITP) grant funded by the Korea government (MSIT) (No.2020-0-00833, A study of 5G based Intelligent IoT Trust Enabler; No. 2022-0-00124, Development of Artificial Intelligence Technology for Self-Improving Competency-Aware Learning Capabilities).
The work of Junsik Kim was supported by Basic Science Research Program through the National Research Foundation of Korea(NRF) funded by the Ministry of Education(2020R1A6A3A01100087).

\bibliographystyle{IEEEtran}
\bibliography{mybib}

\begin{IEEEbiography}[{\includegraphics[width=1in,height=1.25in,clip,keepaspectratio]{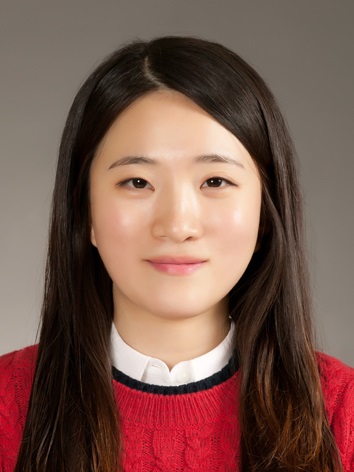}}]{Minkyung Kim} received the B.E. degree in Electrical and Computer Engineering from University of Seoul, South Korea in 2016, and the M.S. and Ph.D. degrees in Electrical Engineering from KAIST, South Korea in 2018 and 2023, respectively. Her research interests include anomaly detection, unsupervised learning, and active learning.
\end{IEEEbiography}

\begin{IEEEbiography}[{\includegraphics[width=1in,height=1.25in,clip,keepaspectratio]{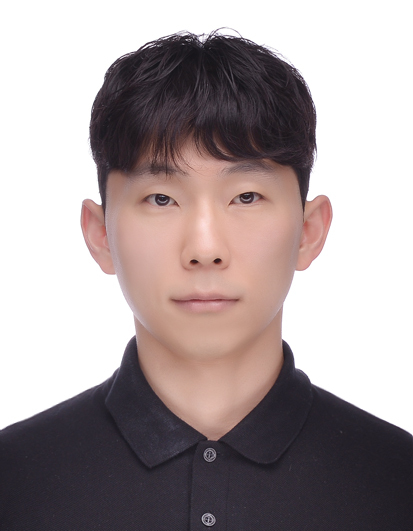}}]{Jongmin Yu} is a postdoctoral research associate at King's College London. He was a research associate at the Institute of IT Convergence at the Korea Advanced Institute of Science and Technology (KAIST). He received PhDs from the School of Electrical Engineering and Computer Science at Gwangju Institute of Science and Technology (GIST), Korea, Republic of, and the School of Electrical Engineering, Computing and Mathematical Sciences at Curtin University, Australia. His research interests include artificial intelligence, machine learning, pattern recognition, and mathematical understanding of these.
\end{IEEEbiography}

\begin{IEEEbiography}[{\includegraphics[width=1in,height=1.25in,clip,keepaspectratio]{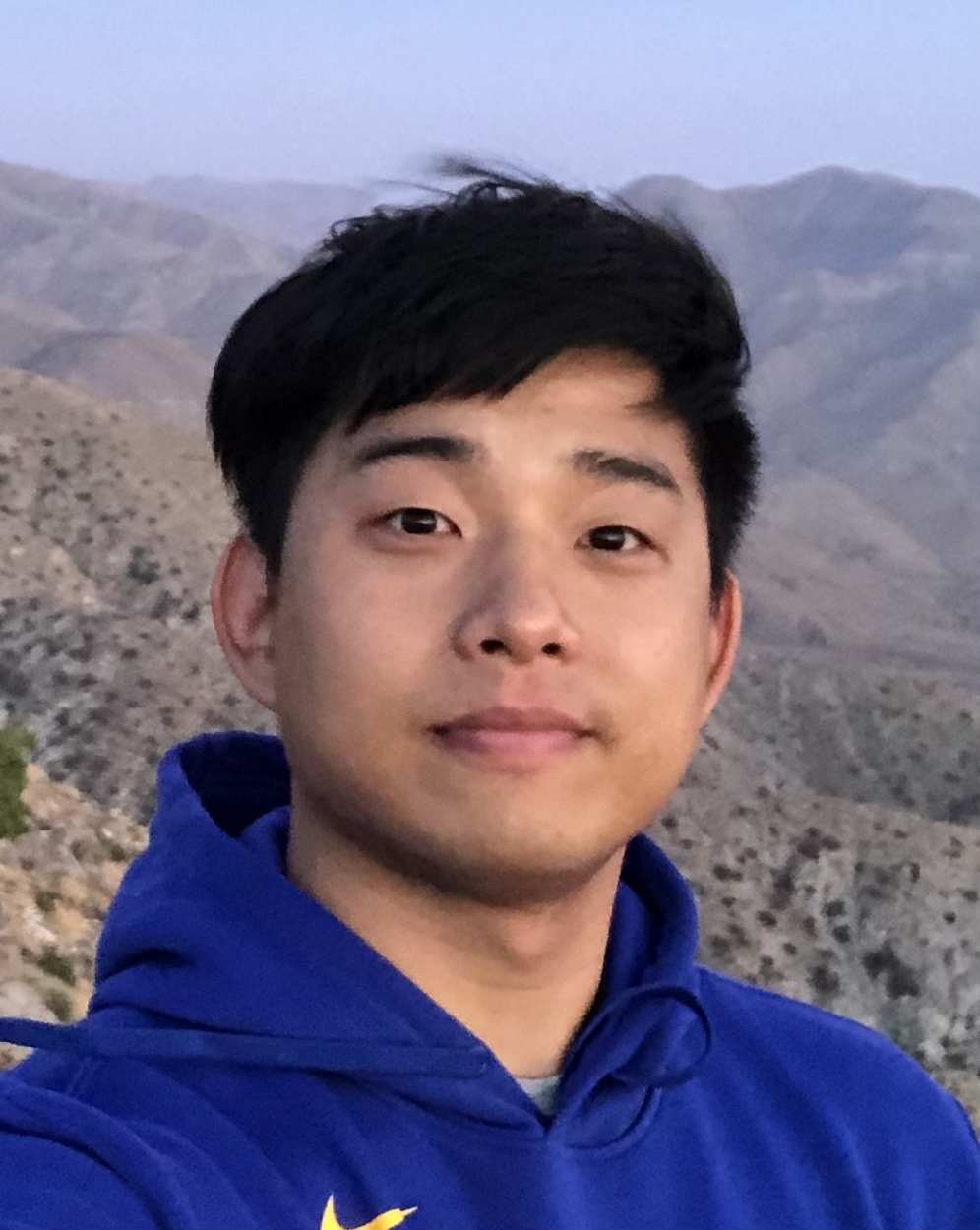}}]{Junsik Kim} received the BS, MS, and Ph.D. degrees in Electrical Engineering Department, KAIST, South Korea, in 2013, 2015, and 2020 respectively. He is currently a postdoctoral researcher in the School of Engineering and Applied Sciences with the Harvard University. Before joining Harvard, he was a postdoctoral researcher with KAIST. His research interest includes computer vision problems, especially with data imbalance and scarcity problems. He was a research intern with Hikvision Research America, Santa Clara, in 2018. He was a recipient of the Qualcomm Innovation award in 2018, and was selected as an outstanding reviewer in ICLR'21.
\end{IEEEbiography}

\begin{IEEEbiography}[{\includegraphics[width=1in,height=1.25in,clip,keepaspectratio]{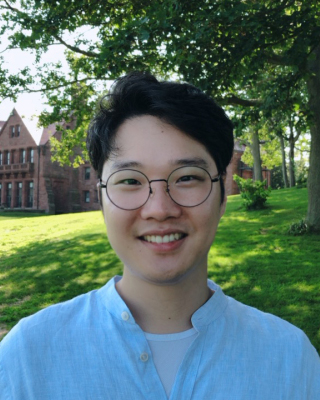}}]{Tae-Hyun Oh}
is an assistant professor with Electrical Engineering (adjunct with Graduate School of AI and Dept. of Creative IT Convergence) at POSTECH, South Korea. He is also a research director at OpenLab, POSCO-RIST, South Korea. 
He received the B.E. degree (First class honors) in Computer Engineering from Kwang-Woon University, South Korea in 2010, and the M.S. and Ph.D. degrees in Electrical Engineering from KAIST, South Korea in 2012 and 2017, respectively.
Before joining POSTECH, he was a postdoctoral associate at MIT CSAIL, Cambridge, MA, US, and was with Facebook AI Research, Cambridge, MA, US. 
He was a research intern at Microsoft Research in 2014 and 2016. He serves as an associate editor for the Visual Computer journal.
He was a recipient of Microsoft Research Asia fellowship, Samsung HumanTech thesis gold award and bronze awards, Qualcomm Innovation awards, and top research achievement awards from KAIST, and was also selected as an outstanding reviewer in CVPR'20 and ICLR'22. 
\end{IEEEbiography}

\begin{IEEEbiography}[{\includegraphics[width=1in,height=1.25in,clip,keepaspectratio]{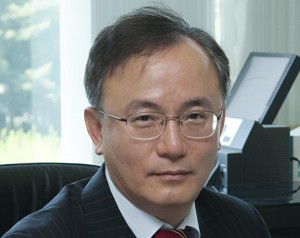}}]{Jun Kyun Choi} received the B.Sc. (Eng.) in Electronics Engineering from Seoul National University, South Korea in 1982, and the M.Sc. (Eng.) and Ph.D. degrees in Electronics Engineering from Korea Advanced Institute of Science and Technology (KAIST), South Korea in 1985 and 1988, respectively.
From 1986 to 1997, he was with the Electronics and Telecommunication Research Institute (ETRI). In 1998, he joined the Information and Communications University (ICU), Daejeon, Korea as Professor, and in 2009, he moved to KAIST as Professor.
He is a Senior Member of IEEE, an executive member of the Institute of Electronics Engineers of Korea (IEEK), an Editor Board Member of the Korea Information Processing Society (KIPS), and a Life member of the Korea Institute of Communication Science (KICS).
\end{IEEEbiography}

\vfill

\end{document}